\documentclass[runningheads]{llncs}

\usepackage{eccv}

\usepackage{eccvabbrv}

\usepackage{graphicx}
\usepackage{booktabs}
\usepackage{multirow}
\usepackage{makecell}
\usepackage{amsmath}
\usepackage{amssymb}
\usepackage{mathtools}
\usepackage{pifont}
\newcommand{\CheckmarkBold}{\ding{51}}
\newcommand{\XSolidBrush}{\ding{55}}

\usepackage{algorithm}
\usepackage{algpseudocode}
\algblock{Loop}{EndLoop}
\algrenewtext{Loop}{\textbf{loop}}
\algrenewtext{EndLoop}{\textbf{end loop}}
\newcommand{\KVSet}{\mathbf{KV}}
\newcommand{\ModelOuput}{\mathbf{X}_{\theta}}
\newcommand{\ModelOuputTemp}{\mathbf{X}_{i}}
\newcommand{\NoisyLatent}{\mathbf{X}_{cache}}
\newcommand{\SinkFrame}{\mathbf{Sink}}
\newcommand{\KVOutput}{\mathbf{kv}}

\usepackage[accsupp]{axessibility}  % Improves PDF readability for those with disabilities.

\usepackage{hyperref}

\usepackage{orcidlink}

\begin{document}

\title{Live Avatar: Streaming Real-time Audio-Driven Avatar Generation with Infinite Length}
\titlerunning{Live Avatar: Streaming Real-time Audio-Driven Avatar Generation}

\author{Yubo Huang\inst{1,2} \and
Hailong Guo\inst{2,3} \and
Fangtai Wu\inst{2,4} \and
Weiqiang Wang\inst{5} \and
Shijie Huang\inst{2} \and
Qijun Gan\inst{4} \and
Shifeng Zhang\inst{2} \and
Lin Liu\inst{1} \and
Sirui Zhao\inst{1}$^*$ \and
Enhong Chen\inst{1}$^*$ \and
Jiaming Liu\inst{2}$^{*\dagger}$ \and
Steven Hoi\inst{2}}

\authorrunning{Y.~Huang et al.}

\institute{University of Science and Technology of China, China \and
Qwen Applications Business Group of Alibaba, China \and
Beijing University of Posts and Telecommunications, China \and
Zhejiang University, China \and
Monash University, Australia\\
\email{siruit@ustc.edu.cn, cheneh@ustc.edu.cn, jmliu1217@gmail.com}\\[2pt]
{\small $^*$Corresponding authors.\quad $^\dagger$Project leader.}}

\maketitle

\begingroup\setlength{\topsep}{0pt}
\begin{abstract}
Audio-driven avatar interaction demands real-time, streaming, and infinite-length generation---capabilities fundamentally at odds with the sequential denoising and long-horizon drift of current diffusion models.
We present Live Avatar, an algorithm-system co-designed framework that addresses both challenges for a 14-billion-parameter diffusion model.
\textbf{On the algorithm side,} a two-stage pipeline distills a pretrained bidirectional model into a causal, few-step streaming one whose KV cache stores \emph{noisy} rather than clean history. This lossy conditioning, counter-intuitively, suppresses error accumulation and enables models trained on short clips to generalize stably beyond 10{,}000 seconds.
\textbf{On the system side,} this \emph{noisy} conditioning also breaks the sequential sampling bottleneck, allowing Timestep-forcing Pipeline Parallelism (TPP) to assign each GPU a fixed denoising timestep, converting the sequential diffusion chain into an asynchronous spatial pipeline that simultaneously boosts throughput and improves temporal consistency.
Live Avatar achieves 45 FPS with a TTFF of 1.21\,s on 5 H100 GPUs, and to our knowledge is the first to enable practical real-time streaming of a 14B diffusion model for infinite-length avatar generation. Our project page is at \url{https://liveavatar.github.io/}.
\end{abstract}\endgroup

\section{Introduction}
\label{sec:intro}
\begin{table}[t]
    \centering
    \small
    \caption{Comparison of state-of-the-art audio-driven avatar generation methods. Live Avatar simultaneously achieves streaming, real-time, and infinite-length generation with a large-scale (14B) diffusion model.}
    \label{tab:method_comparison}
    \setlength{\tabcolsep}{0pt}
    \begin{tabular*}{0.8\columnwidth}{@{\extracolsep{\fill}}l|cccc}
        \toprule
        \textbf{Method} & streaming & real-time & infinite-length & scale \\
        \midrule
Hallo3\cite{cui2025hallo3} & \XSolidBrush & \XSolidBrush & \XSolidBrush & 5B  \\
StableAvatar\cite{tu2025stableavatar} & \XSolidBrush & \XSolidBrush & \CheckmarkBold & 1.3B  \\
Wan-S2V\cite{gao2025wans2vaudiodrivencinematicvideo} & \XSolidBrush & \XSolidBrush & \XSolidBrush & 14B \\
Ditto\cite{li2024ditto} & \CheckmarkBold & \CheckmarkBold & \CheckmarkBold & 0.2B \\
InfiniteTalk\cite{yang2025infinitetalk} & \XSolidBrush & \XSolidBrush & \CheckmarkBold & 14B \\
OmniAvatar\cite{gan2025omniavatar} & \XSolidBrush & \XSolidBrush & \XSolidBrush & 14B \\
\midrule
Live Avatar (Ours) & \CheckmarkBold & \CheckmarkBold & \CheckmarkBold & 14B \\
\bottomrule
    \end{tabular*}
\end{table}
Audio-driven avatar generation, the synthesis of photorealistic human face video whose motion is driven by an input audio stream, is a foundational technology for interactive digital communication. Its applications are expansive, ranging from virtual reality and live streaming to digital assistants. The demand for systems capable of producing high-fidelity, expressive, and real-time avatars has driven significant recent advancements, particularly with the rise of diffusion models for high-fidelity video synthesis~\cite{videoworldsimulators2024,hacohen2024ltx,wan2025wan}.
Despite their success in setting new benchmarks for visual quality, deploying these powerful generative models in real-time\footnote{Throughout this paper, ``real-time'' refers to generation throughput exceeding playback speed, not end-to-end conversational latency.} ($\geq$\,24\,FPS), streaming environments over unbounded durations faces fundamental and conflicting challenges.

The first challenge is long-horizon consistency. Existing methods suffer from compounding errors that cause identity drift and color artifacts within minutes~\cite{tu2025stableavatar}. Recent progress such as Self-Forcing~\cite{huang2025self} mitigates the train--test gap but still degrades rapidly under minute-level rollout; LongLive~\cite{yang2025longlive} sustains longer generation via native long-duration training but is designed for text-to-video and too costly to scale to 14B models.

The second challenge is the real-time--fidelity trade-off. Large-scale diffusion models~\cite{gao2025wans2vaudiodrivencinematicvideo} yield superior visual quality but are inherently slow due to sequential multi-step denoising, while existing real-time approaches~\cite{li2024ditto} sidestep this sequential bottleneck entirely by adopting non-iterative generation methods or very small models, at the cost of fidelity. Reconciling these two desiderata at scale remains an open problem.

% To address this critical challenges, we propose \textbf{Live Avatar}, a novel training and inference framework. Live Avatar is designed to fundamentally enable large diffusion models (up to 14 Billion parameters) for real-time, streaming, and infinite-length audio-driven avatar generation without compromising visual fidelity. Our work successfully resolves the long-standing tension between high visual quality, model complexity, and practical execution speed through an algorithm-system co-design approach.
To address these critical challenges, we propose \textbf{Live Avatar}, an algorithm-system co-designed framework that enables large diffusion models (up to 14 billion parameters) for real-time, streaming, and infinite-length audio-driven avatar generation without compromising visual fidelity. 

For infinite-length stability, we leverage the static-scene prior of avatar interaction to anchor identity in a persistent sink frame and store only \emph{noisy} representations in the KV cache---the noise acts as a low-pass filter that suppresses accumulated artifacts while preserving motion dynamics, and keeps the latent distribution compact to prevent out-of-distribution drift. Additional mechanisms further address distribution drift and positional extrapolation, together enabling stable generation beyond 10{,}000 seconds.

For real-time streaming, we distill the model into a few-step causal one via Self-Forcing and propose Timestep-forcing Pipeline Parallelism (TPP), which exploits the \emph{noisy} KV cache to pipeline denoising steps across GPUs, breaking the sequential sampling bottleneck. Combined with system-level optimizations, this achieves high throughput and low latency at 14B scale.

The core contributions of Live Avatar are as follows:

\begin{itemize}
    \item \textbf{Causal, Streamable Adaptation Framework.} We propose a two-stage framework that adapts a pretrained bidirectional diffusion model into a causal, few-step streaming model, with a novel motion-frame-as-scaffold mechanism that bridges Stage~1 and Stage~2 by providing functionally analogous training signals, yielding a $5\times$ distillation convergence speedup.

    \item \textbf{Long-Horizon Stability.} We show that storing only \emph{noisy} history in the KV cache enables train-short-infer-long: training on ${\sim}$3\,s clips yet extrapolating stably beyond 10{,}000\,s. Additional mechanisms further address distribution drift and positional extrapolation.

    \item \textbf{Timestep-forcing Pipeline Parallelism (TPP).} We propose TPP, which exploits the \emph{noisy} KV cache to assign each GPU a fixed denoising timestep, breaking the sequential sampling bottleneck and boosting throughput. Combined with system-level optimizations, TPP achieves 45 FPS with a TTFF of 1.21\,s on 5$\times$H100 GPUs, to our knowledge the first practical real-time streaming of a 14B diffusion model.

    \item \textbf{Standardized Benchmark.} We introduce GenBench (Short/Long), including long-video test cases exceeding five minutes.
% a cluster of 
\end{itemize}

%Leveraging these innovations, Live Avatar achieves a significant performance breakthrough: to the best of our knowledge, we are the first to demonstrate stable real-time streaming of a 14-billion-parameter model at an end-to-end 20 FPS using multi-step (4-step) denoising. This result substantially surpasses existing baselines and paves the way for the industrialization and real-time deployment of high-fidelity video diffusion models.

\section{Related Work}
\noindent\textbf{Streaming and Long Video Generation.} Streaming and long video generation require efficient management of both computation and memory resources~\cite{yin2025slow,kodaira2025streamdit,liu2025rolling,cui2025self}. Diffusion Forcing~\cite{chen2024diffusion} introduces varied noise levels to sequential targets, enabling flexible-length streaming generation. Self-Forcing~\cite{huang2025self} addresses train-inference mismatch by conditioning on previously generated frames, yet still suffers from exposure bias and fails to produce minute-level long videos. LongLive~\cite{yang2025longlive} applies sliding window distillation for streaming long video generation, but is natively designed for text-to-video and unsuitable for image-conditioned tasks, and its training inefficiency prevents scaling to large-scale models such as 14B-parameter architectures. While these approaches improve video quality and temporal consistency, none achieve real-time, streaming long video generation with large-scale diffusion models.

\noindent\textbf{Audio-driven Avatar Video Generation.}
Audio-driven avatar video generation requires subject consistency and effective motion control. Early works~\cite{prajwal2020lip,zhang2023sadtalker} leverage GANs and 3D motion models for lip-syncing and facial animation. With the success of diffusion models, several studies~\cite{tian2024emo,yu2025llia} adapt diffusion frameworks and ReferenceNet architectures for avatar generation. DiT-based video generation models~\cite{videoworldsimulators2024,hacohen2024ltx,kong2024hunyuanvideo,wan2025wan} have demonstrated remarkable visual quality, inspiring a surge of DiT-based avatar methods~\cite{du2025rap, meng2025mirrorme, low2025talkingmachines,li2024ditto, wang2025omnitalker, zhen2025teller, guo2025arig, zhu2025infp}. Meanwhile, autoregressive approaches~\cite{li2024autoregressive,ao2024body,chen2025midas,xie2025x} offer an alternative by combining autoregressive and diffusion strategies. However, none achieve real-time, streaming, infinite-length avatar generation with large-scale diffusion models.

\noindent\textbf{Diffusion Distillation.}
Diffusion model distillation accelerates video generation through various paradigms, including trajectory distillation~\cite{frans2024one}, Consistency Models~\cite{song2023consistency,zheng2025large,lu2024simplifying}, and Distribution Matching Distillation (DMD)~\cite{yin2024one,luo2025learning,luo2023diff}. Among these, DMD has proven particularly effective for streaming video generation: CausVid~\cite{yin2025slow}, Self-Forcing~\cite{cui2025self,huang2025self}, and LongLive~\cite{yang2025longlive} all employ DMD-based few-step distillation on temporally segmented long videos, achieving both significant sampling speedup and improved long-video quality. Recent analysis further suggests that DMD functions analogously to reinforcement learning, with the pretrained diffusion model serving as a reward signal~\cite{luo2024diff}, potentially explaining its effectiveness in streaming settings.
% \section{Method}
\section{Preliminaries}
\label{sec:preliminaries}
\subsection{Video Diffusion Models}
Video diffusion models generate high-fidelity video sequences by progressively denoising from a Gaussian prior $x_T\sim N(0,I)$, following the reverse process of a forward diffusion. In this work, we adopt the flow matching~\cite{lipman2022flow}, where noisy latents at time $t$ are constructed as

\begin{equation}
x_t = (1 - s_t) \cdot x_0 +s_t \cdot x_T,
\end{equation}
where $s_t\in [0,1]$ is a scheduling function that controls the interpolation between clean and noise. The model is trained to predict the target velocity
\begin{equation}
v=x_T-x_0
\end{equation}
leading to the standard mean squared error objective:
 \begin{equation}
\mathcal{L} = \mathbb{E}_{x_0, x_T, t} \left[ \| v_\theta(x_t, t, c) - (x_T - x_0) \|_2^2 \right]
\end{equation}
where $c$ denotes conditional inputs such as text embeddings.

To reduce computational cost, most modern video diffusion models operate in a compressed latent space. We use a causal 3D VAE to encode input videos into temporal-latent representations, ensuring that future frames do not leak during training. Text conditioning is achieved through a pre-trained language model that produces contextual embeddings fed into the diffusion backbone.

% The architecture of choice is DiT (Diffusion Transformer)~\cite{peebles2023scalable}, which replaces U-Net convolutions with transformer blocks equipped with factorized spatial-temporal attention. This design enables better scalability and long-range modeling. 

\subsection{Distribution Matching Distillation}
Distribution Matching Distillation (DMD)~\cite{yin2024one} aims to distill a pre-trained teacher diffusion model into a student model that operates with fewer sampling steps.  Let \( p_{\theta,t}(\mathbf{x}_t) \) denote the distribution induced by the few-step student model \( \mathbf{x} = G_\theta(\mathbf{z}) \), and let \( p_{\text{data},t}(\mathbf{x}_t) \) represent the corresponding ground-truth distribution produced by the teacher diffusion model at time step \( t \). The primary objective of DMD is to minimize the distribution, i.e., reverse Kullback--Leibler (KL) divergence, between these two distributions at each time step \( t \): $\mathbb{E}_t \left[ D_{\mathrm{KL}} \big( p_{\theta,t} \,\|\, p_{\text{data},t} \big) \right].$ The gradient of DMD loss is given by:
\begin{equation}
\nabla_\theta \mathcal{L}_{\mathrm{DMD}} = - \mathbb{E}_{t, \mathbf{z}} \left[ \left( s_{\text{real}}(\mathbf{x}_t, t) - s_{\text{fake},\phi}(\mathbf{x}_t, t) \right)^\top \frac{\partial G_\theta(\mathbf{z})}{\partial \theta} \right]
\label{equ:dmd2}
\end{equation}
where \(\mathbf{x}_t = \Psi(\hat{\mathbf{x}}, t) \) is the noise scheduler, \( \hat{\mathbf{x}} = G_\theta(\mathbf{z}) \) is the data prediction of the distilled model, \( s_{\text{real}} \) and \( s_{\text{fake},\phi} \) denote the score functions corresponding to the pre-trained teacher diffusion model and the student generator, respectively.

\( s_{\text{real}} \), \( s_{\text{fake},\phi} \), and \( G_\theta(\mathbf{z}) \) are all initialized from the pre-trained teacher model induced by $v_\theta$. The training proceeds by alternately updating \( s_{\text{fake},\phi} \) and \( G_\theta(\mathbf{z}) \), in which \( s_{\text{fake},\phi} \) is trained on samples generated by the current student generator \( G_\theta(\mathbf{z}) \), and \( G_\theta(\mathbf{z}) \) is trained using the DMD loss defined above. For multi-step distillation, DMD first performs a multi-step sampling trajectory using the student generator: $\mathbf{z} \xrightarrow{G_\theta} \hat{\mathbf{x}}_{t_1} \xrightarrow{\Psi} \mathbf{x}_{t_1} \xrightarrow{G_\theta} \hat{\mathbf{x}}_{t_2} \xrightarrow{\Psi} \mathbf{x}_{t_2} \xrightarrow{} \cdots \xrightarrow{} \mathbf{x}_{t_N}.$
Then, at each training iteration, a random intermediate state \( \mathbf{x}_{t_i} \) from this trajectory is selected and used in place of pure noise \( \mathbf{z} \) as the starting point for the DMD training procedure~\cite{yin2024improved}.
\begin{figure}[t]
  \centering
  % \fbox{\rule{0pt}{2in} \rule{0.98\linewidth}{0pt}}
   \includegraphics[width=0.98\linewidth]{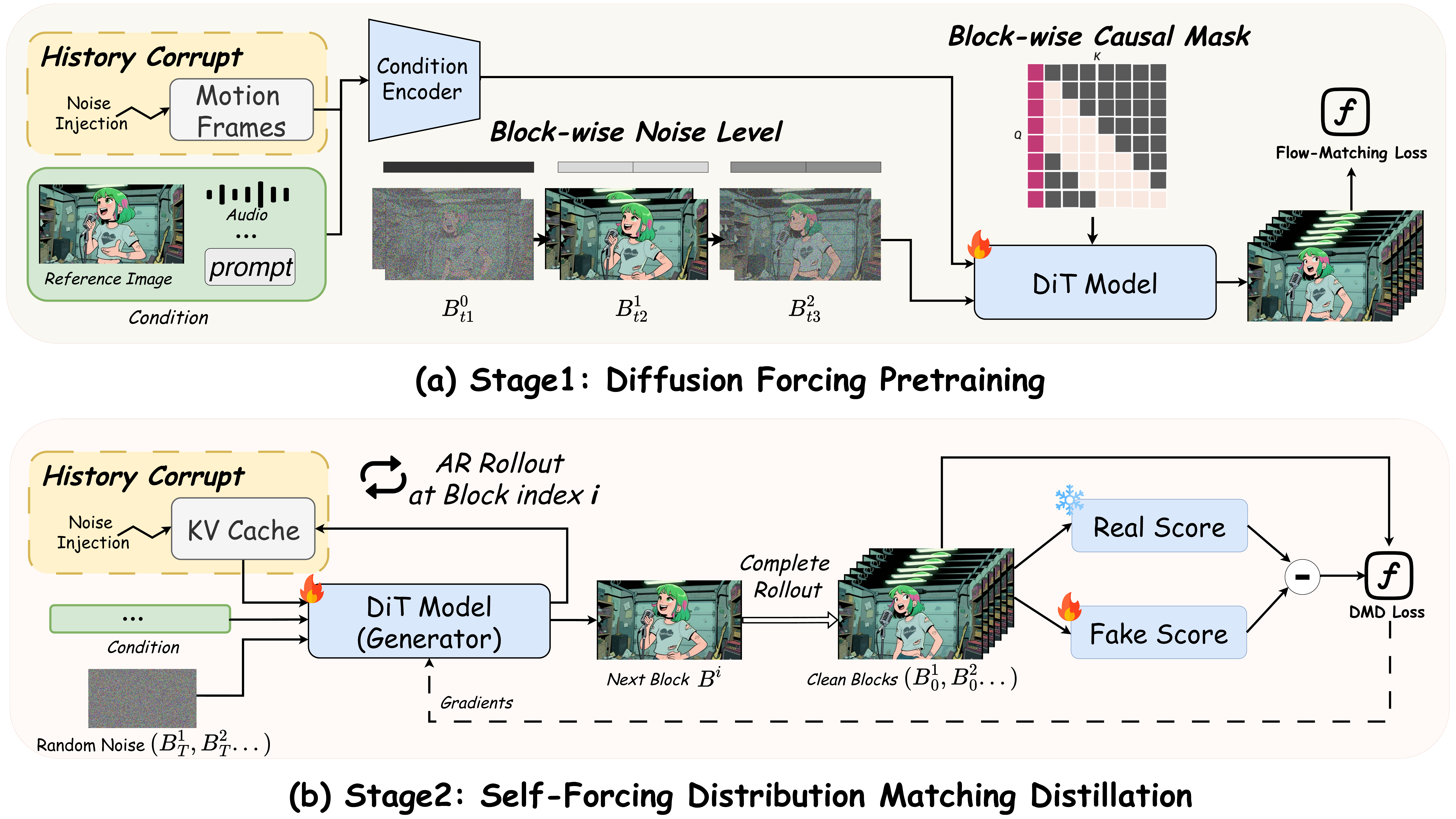}

   \caption{The Live Avatar Training Framework. (a) Stage 1 Diffusion Forcing Pretraining with motion-frame scaffolding: noisy motion frames provide auxiliary temporal context alongside block-wise independent noise and causal attention masks. (b) Stage 2 Self-Forcing Distillation with History Corrupt: the motion frames are removed and replaced by the rolling KV cache, where noise level consistency between the KV cache and the noisy latents is enforced.}
   \label{fig:framework_train}
\end{figure}

\section{The Live Avatar Framework}

In this section, we present the technical details of Live Avatar. We first detail the model architecture in Sec.~\ref{sec:architecture}, followed by the overall training framework in Sec.~\ref{sec:training} and Fig.~\ref{fig:framework_train}. We then investigate long video generation in Sec.~\ref{sec:rolling_anchor}, proposing three strategies to address visual quality degradation, identity drift, and color artifacts in long-term generation. Finally, the inference framework and Timestep-forcing Pipeline Parallelism are demonstrated in Sec.~\ref{sec:inference}.

\subsection{Model Architecture}
\label{sec:architecture}
In order to enable streaming video generation, the Live Avatar adopts autoregressive generation by factorizing the joint distribution 
\begin{equation}
\label{equ:joint}
B_{t-1}^{i} = D_\theta(B_t^i, \underbrace{B_t^{(i-w):(i-1)}}_{\text{kv cache}}, I, a^i, t^i)
\end{equation}
combining diffusion-based frame synthesis with causal dependencies across chunks. $B$ in Eq.~\ref{equ:joint} are blocks of consecutive noisy frame latents. In our work, we set the number of frame latents to 3. $w$ is the rolling KV-cache window size. $I$ denotes the sink frame, which provides the appearance; $a^i$ and $t^i$ are the audio embedding and prompt embedding for the $i$-th block respectively. The underscore $t$ denotes the denoising step, and the superscript $i$ denotes the block index. Note that in the model, the kv cache $B_t^{(i-w):(i-1)}$ and the noisy block $B_t^i$ share the same noise level, this is a crucial design that improves generation quality and maximize the inference speed, which will be illustrated in Sec.~\ref{sec:inference}.

\subsection{Model Training}
\label{sec:training}
% Training
Our overall training framework is illustrated in Figure~\ref{fig:framework_train}, which consists of two stages: 1) Stage 1, \textbf{Diffusion Forcing Pretraining} with motion-frame scaffolding, and 2) Stage 2, \textbf{Self-Forcing Distribution Matching Distillation} with History Corrupt.

\noindent\textbf{Stage 1: Diffusion Forcing Pretraining.}
Following prior practice~\cite{chen2024diffusion,yin2025slow}, we apply block-wise independent noise scheduling and causal attention masks---frames within a block share full attention while inter-block attention is strictly causal---and train the model with the standard flow-matching loss. We additionally introduce motion frames~\cite{cui2025hallo3} as auxiliary context, whose role is detailed below.

\noindent\textbf{Motion-Frame as Scaffold.}
Employing a naive two-stage training converges slowly on long-horizon capabilities, yet each self-forcing rollout step is far more expensive than supervised training. Inspired by~\cite{gelberg2025extending}, we repurpose the motion frames---context frames preceding the current clip, originally for clip continuation---as scaffolding in Stage~1 by injecting noise into them. Since both noisy motion frames and the Stage~2 noisy KV cache serve a functionally analogous role as temporal context for continuing generation, this cheaply teaches dynamics--identity decoupling without costly self-forcing rollouts. The motion frames are then entirely replaced by the KV cache in Stage~2, yielding a $5\times$ convergence speedup.

\noindent\textbf{Stage 2: Self-Forcing Distillation.}
In Stage~2 we distill the bidirectional teacher into a causal, few-step streaming model following Self-Forcing~\cite{huang2025self}. The causal student denoises one block at a time while conditioning on a rolling KV cache of previously generated blocks.
Crucially, we omit the extra clean-cache refresh forward pass used in prior work~\cite{huang2025self}, so that the KV cache always contains \emph{noisy} representations---a strategy we call \textbf{History Corrupt}, whose motivation is detailed in Sec.~\ref{sec:rolling_anchor}.
During the denoising of each block, the model attends to the KV cache from previous blocks at the \emph{same} timestep; the noise-level implications of this design are also discussed in Sec.~\ref{sec:rolling_anchor}.

% ------

% 推理

% 我们的推理流程图如\ref{fig:framework_infer}所示，核心思想为timestep-based分布式推理。具体来说，我们为每个gpu设备分配一个timestep，每个设备只会负责一个timestep的降噪, 每个设备会提前获取audio和文本的features然后释放对应encoder。因为降噪是从纯噪声开始，所以靠后的timestep的设备需要等靠前timestep的设备先完成降噪才能启动自己的降噪，因为降噪步数是常数的所以延后时间也是常数，对利用率影响非常小。如图所示，从第一个block即B_1的纯噪声时间步t_T开始，由GPU0完成一步降噪后首先将下一时间t_{s_3}的noisy latents传递给下一设备。然后GPU0开始下一个block的降噪，下一个block同样是GPU0从纯噪声开始降噪，不同的是这次降噪应用了rollout KV cache，可以看到前面多个block到KV cache特征，这对于维持时间一致性是必要的，rollout意味着KV cache池子大小有上限比如限制token数不超过N，达到限制时过于久远的block的KV cache会被扔掉。我们同时查看GPU1的进程,GPU1一但接受到B_1的t_{s_3}的noisy latents，就会立刻开始降噪得到t_{s_2}的noisy latents传递给GPU2并且自己也会接受来自GPU0的t_{s_3}的属于B_2的noisy latents开始利用B_1的cache KV进行B2对t_{s_3}的降噪，后面如法炮制。到最后一步，由一个独立设备负责将t_0的latents通过3D VAE decode得到block-by-block的视频片段并音频组合起来得到最终的视频。

% \begin{algorithm}
% \caption{Timestep-forcing Pipeline Parallelism}
% \begin{algorithmic}[1]
% \Procedure{QuickSort}{$A, p, r$}
%     \If{$p < r$}
%         \State $q \gets \text{Partition}(A, p, r)$
%         \State \Call{QuickSort}{$A, p, q-1$}
%         \State \Call{QuickSort}{$A, q+1, r$}
%     \EndIf
% \EndProcedure

% \Procedure{Partition}{$A, p, r$}
%     \State $x \gets A[r]$
%     \State $i \gets p - 1$
%     \For{$j = p$ to $r - 1$}
%         \If{$A[j] \leq x$}
%             \State $i \gets i + 1$
%             \State swap $A[i]$ and $A[j]$
%         \EndIf
%     \EndFor
%     \State swap $A[i+1]$ and $A[r]$
%     \State \Return $i + 1$
% \EndProcedure
% \end{algorithmic}
% \end{algorithm}

% Inference

% \subsubsection{Rethinking Reference Conditioning}
\subsection{Long Video Generation}
\label{sec:rolling_anchor}

Existing talking-avatar systems exhibit pronounced degradation over long, autoregressive generation---manifesting as identity drift, color shifts, and temporal instability~\cite{tu2025stableavatar}. 
In practice, we perform inference in a rolling KV cache fashion~\cite{huang2025self}, which extends the temporal horizon but, on its own, does not prevent collapse.
We attribute these long-horizon failures to three internal phenomena:

\noindent\textbf{(i) Test-time conditioning drift.} The conditioning pattern at test time (e.g., the RoPE-relative positioning between the sink frame and current target blocks) gradually diverges from the training-time setup, weakening identity cues.

\noindent\textbf{(ii) Distribution drift.} The distribution of generated frames progressively deviates from realistic video distributions, likely driven by persistent factors (e.g., a real-data sink frame whose distribution subtly differs from the model's generation manifold) that continuously push the rolling generation toward unrealistic outputs.

\noindent\textbf{(iii) Error accumulation.} Subtle imperfections in each generated block are inherited and compounded frame-by-frame through the clean KV cache, as the model attends to fine-grained details---including artifacts---from previous blocks. This compounding causes rapid quality deterioration over time.

To address these challenges, we propose three complementary strategies---History Corrupt, Adaptive Attention Sink, and Rolling RoPE---that together enable stable infinite-length generation.

\noindent\textbf{History Corrupt.}
For avatar interaction, the subject typically resides in a relatively static scene: apart from facial expressions, body gestures, and mild background dynamics, the visual content does not undergo rapid change. This prior allows a simplifying design assumption---the sink frame can provide sufficiently useful appearance and identity information for \emph{every} generated frame throughout the entire sequence. Under this assumption, the role of the rolling KV cache reduces to conveying \emph{motion dynamics} alone, while fine-grained identity and appearance details should be sourced exclusively from the persistent sink frame.

This motivates a simple yet effective design: we store only \emph{noisy} representations in the KV cache, rather than the conventional clean cache.
Intuitively, Gaussian noise acts as a low-pass filter~\cite{song2025historyguidedvideodiffusion} on the cached representations, suppressing high-frequency details (including accumulated artifacts) while preserving low-frequency motion structure. This forces the model to extract dynamic cues from the noisy history while relying on the clean sink frame for identity and appearance, achieving an effective \emph{dynamics--identity decoupling}. As a result, generation errors in previous blocks are no longer faithfully propagated, directly addressing error accumulation~(iii).

Furthermore, noisy KV caches also mitigate distribution drift~(ii). At higher noise levels, the marginal distribution $p_t(\mathbf{x}_t)$ converges toward the Gaussian prior and occupies a progressively more compact region of the latent space~\cite{song2019generative}. Consequently, the noisy cache is far less likely to drift into out-of-distribution regions compared to a clean cache, which must remain on the narrow data manifold where small perturbations can compound into distributional shift.

We further observe empirically that enforcing the noisy block and its attended KV cache to share the \emph{same} noise level---a strategy we call \textbf{timestep-forcing}---significantly reduces inter-frame flickering, consistent with observations in TalkingMachines~\cite{low2025talkingmachines}. This timestep-forcing constraint also naturally lends itself to efficient multi-GPU inference, as discussed in Sec.~\ref{sec:inference}.

\noindent\textbf{Adaptive Attention Sink (AAS)\footnote{Details of AAS and Rolling RoPE are provided in the supplementary materials.}.}
By default, the user-provided reference image (i.e., the image-to-video conditioning input) serves as the sink frame. However, this real-data image resides on a slightly different distribution from the model's own generation manifold, introducing a persistent bias that accumulates into color, exposure, or style deviations over long runs.
To counteract this form of distribution drift~(ii), AAS replaces the sink frame with the model's own first generated latent immediately after the first block is produced, and uses it as the persistent sink for all subsequent conditioning. By keeping the sink frame within the model's learned generation manifold, AAS eliminates the distributional mismatch between conditioning and generated content.

\noindent\textbf{Rolling RoPE\footnotemark[\value{footnote}].}
To mitigate test-time conditioning drift~(i), we introduce a dynamic position-alignment mechanism for the sink frame. The sink frame is permanently cached in KV memory and its temporal offset is adjusted via a controllable RoPE shift so that its relative position to the current noisy states remains consistent with training. This dynamic RoPE alignment lets the model continuously reference identity features from the sink frame without rigidly constraining local motion, thereby stabilizing long-range identity and structural fidelity.

\begin{figure}[t]
  \centering
  % \fbox{\rule{0pt}{2in} \rule{0.98\linewidth}{0pt}}
   \includegraphics[width=0.98\linewidth]{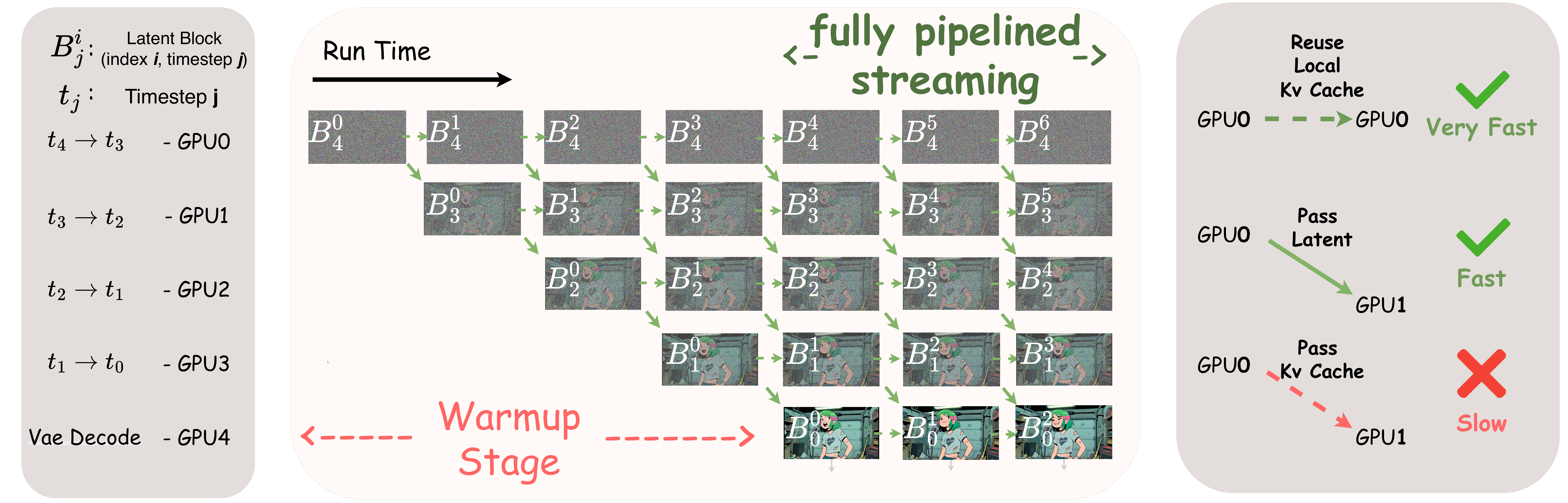}

   \caption{A visual illustration of Timestep-forcing Pipeline Parallelism (\textbf{TPP}). After warm-up fills the pipeline, all GPUs denoise \textbf{simultaneously} in the fully pipelined stage, turning the sequential diffusion chain into an asynchronous spatial pipeline. For example, GPU2 always performs the $t_2 \rightarrow t_1$ step: it reuses its local KV cache (\textbf{very fast}) and sends only the latent to GPU3 (\textbf{fast}).}
   \label{fig:framework_infer}
\end{figure}

\subsection{Timestep-forcing Pipeline Parallelism}
\label{sec:inference}

Deploying large video generation models in real-time settings remains challenging due to the inherent sequential structure of diffusion-based sampling. Our distilled 14B model, for example, reaches only 5 FPS on a single GPU and 6 FPS under conventional 4-GPU sequential parallelization. In contrast, existing real-time streaming methods, such as CausVid~\cite{yin2025slow} and LongLive~\cite{yang2025longlive}, achieve higher frame rates by substantially reducing model capacity---often relying on lightweight 1.3B models and aggressive quantization---at the cost of generation fidelity. This establishes a long-standing tradeoff between model quality and real-time performance that has yet to be resolved.

To overcome the sequential bottleneck of diffusion sampling, we introduce \textbf{Timestep-forcing Pipeline Parallelism (TPP)} (illustrated in Figure~\ref{fig:framework_infer}), which assigns each GPU a fixed timestep $t_i$ and partitions the $T$ denoising steps across $T$ devices. Each GPU repeatedly performs its designated transformation $t_i$ $\rightarrow$ $t_{i-1}$, converting the sequential diffusion chain into an asynchronous spatial pipeline. Through this reparameterization, throughput is determined by a single denoise forward rather than the sum of all diffusion steps, yielding an ideal speedup proportional to the total number of denoising steps. Importantly, TPP is both \emph{model-agnostic} and \emph{hardware-agnostic}: it applies to any causal diffusion model---not only distilled ones---and only requires that each device can execute a single denoising step with minimal inter-device bandwidth, since TPP communicates only the compact latent between stages.

TPP operates in two stages. During warm-up, the first block is propagated through all timesteps to fill the pipeline, which completes quickly given the small number of sampling steps. Once filled, the system enters the fully pipelined streaming phase: each GPU repeatedly performs its assigned denoising step, passes the latent to the next device, and immediately processes the next block, achieving maximal parallel throughput.
Since TPP assigns one timestep per GPU, the timestep-forcing strategy (Sec.~\ref{sec:rolling_anchor}) maps directly onto the pipeline: each $GPU_i$ maintains its own local rolling KV cache, requiring no inter-GPU communication and thus remaining extremely fast and efficient.
After finishing its step, $GPU_i$ passes only the latent to $GPU_{i+1}$, keeping the communication cost negligible. 
To prevent pipeline bottlenecks, the VAE decoding stage is offloaded to an additional dedicated GPU, which consumes the clean latent and outputs synchronized video chunks.

% ------
\section{Experiments}
\subsection{Experimental Settings}
\noindent
\textbf{Implementation Details.}
The overall model architecture is borrowed from WanS2V~\cite{gao2025wans2vaudiodrivencinematicvideo}. In Stage~1, we initialize from its weights and implement motion frames with the frozen FramePack encoder from WanS2V; each motion frame is independently noised via the flow-matching forward process~\cite{lipman2022flow} with $t\!\sim\!\mathcal{U}[0,1000]$. In Stage~2, the motion-frame encoder is removed, the teacher score and fake score branches are initialized from WanS2V, and the student is initialized from Stage~1.
All training and inference are performed at a fixed resolution of 720$\times$400 and 84 frames. Experiments are conducted on 128 NVIDIA H100 GPUs, with 25\,K steps for Stage~1 and 500 steps for Stage~2. 
The per-GPU batch size is 1. To handle the high memory demand of Self-Forcing training, we adopt FSDP with gradient accumulation to reduce memory consumption. 
The learning rate is 1e-5 for the student and 2e-6 for the fake score. We group every 3 latents into a block, set the KV-cache window size $w\!=\!4$ blocks (Eq.~\ref{equ:joint}), and use a single sink frame. We train models with a LoRA, whose rank and alpha are set to 128 and 64, respectively. 
At inference time, we apply a series of kernel-level optimizations (collectively referred to as Kernel Opt.\ in Table~\ref{tab:Ablation_eff}): FP8 quantization, FlashAttention-3, cuDNN fused kernels, torch.compile, LoRA weight merging, and VAE feature caching which caches intermediate features to enable streaming decoding. All reported metrics reflect this optimized configuration.
% 我们build模型基于Wan-S2V /cite{WanS2V},一个强大的14b的音频到视频生成模型，我们在stage1中从其权重初始化，在stage2中用其为teacher score和fake score初始化，而用stage1的权重为stage2的student初始化。在所有训练和推理过程中，我们限制分辨率为720*400。训练过程所有模型的帧数都设为84.所有实验在128张英伟达H800上进行，stage1和stage2上分别进行了25000step和2500step，总共使用了大约500GPUDays。每卡batchsize都为1。考虑到Self-Forcing训练对显存的极度高需求，我们使用了FSDP，并将单次过程拆解为多个梯度累积步骤来节省显存。在两阶段训练中student的学习率都固定为1e-5，fake score学习率固定为2e-6.我们将连续的每3帧设为一个block，kv cache的长度设置为4个block，rolling sink frame为单独的一帧。lora rank为128，alpha为64.

\noindent
\textbf{Datasets.} We train on AVSpeech~\cite{AVSpeech}, a large-scale audio-visual dataset of talking-head clips. We adopt the preprocessing of OmniAvatar~\cite{gan2025omniavatar} and keep only clips longer than 10\,s, yielding 400K training samples.
To evaluate our model's out-of-domain (OOD) generalization, we created a synthetic benchmark named GenBench. This test set was generated using Gemini-2.5 Pro, Qwen-Image~\cite{wu2025qwenimagetechnicalreport}, and CosyVoice~\cite{du2024cosyvoice2scalablestreaming}. It is composed of two subsets: GenBench-ShortVideo, comprising 100 test samples with an approximate duration of 10 seconds, and GenBench-LongVideo, which contains 15 test videos, each exceeding 5 minutes in duration. The benchmark is designed to be challenging, featuring a wide diversity of character styles (photorealistic humans, animated characters, and anthropomorphic non-humans) and visual compositions, including frontal and profile views, as well as half-body and full-body shots. This variety allows for a robust assessment of the model's performance on unseen data.

\begin{table}[htbp]
\centering
\small
\caption{Quantitative comparisons of our methods with state-of-the-art methods.}
\label{tab:quantitative_comparison}
\resizebox{\columnwidth}{!}{%
\begin{tabular}{l l c c c c lc }
\toprule
\multicolumn{1}{c}{Dataset} & \multicolumn{1}{c}{Model} & \multicolumn{6}{c}{Metrics} \\
\cmidrule(lr){3-8}
 & & ASE $\uparrow$ & IQA $\uparrow$ & Sync-C$\uparrow$ & Sync-D$\downarrow$  &Dino-S $\uparrow$& FPS $\uparrow$ \\
\midrule
\multirow{7}{*}{GenBench-ShortVideo}
 & Ditto\cite{li2024ditto} & 3.31 & 4.24 & 4.09 & 10.76  &\textbf{0.99} & 21.80 \\
 & Echomimic-V2\cite{meng2025echomimicv2strikingsimplifiedsemibody} & 2.82 & 3.61 & 5.57 & 9.13  &0.79 & 0.53 \\
 & Hallo3\cite{cui2025hallo3} & 3.12 & 3.97 & 4.74 & 10.19  &0.94 & 0.26 \\
 & StableAvatar\cite{tu2025stableavatar} & 3.52 & 4.47 & 3.42 & 11.33  &0.93 & 0.64 \\
 & OmniAvatar\cite{gan2025omniavatar} & \textbf{3.53} & 4.49 & 6.77 & \textbf{8.22}  &0.95 & 0.16 \\
 & WanS2V\cite{gao2025wans2vaudiodrivencinematicvideo} & 3.36 & 4.29 & 5.89 & 9.08  &0.95 & 0.25 \\
 & Ours & 3.44 & \textbf{4.51} & \textbf{7.03} & 8.30  &0.96 & \textbf{45.2} \\
\midrule
\multirow{6}{*}{GenBench-LongVideo} 
 & Ditto\cite{li2024ditto} & 2.90 & 4.48 & 3.98 & 10.57  &\textbf{0.97} & 21.80 \\
 & Hallo3\cite{cui2025hallo3} & 2.65 & 4.04 & 6.18 & 9.29  &0.83 & 0.26 \\
 & StableAvatar\cite{tu2025stableavatar} & 3.00 & 4.66 & 1.97 & 13.57  &0.94 & 0.64 \\
 & OmniAvatar\cite{gan2025omniavatar} & 2.36 & 2.86 & \textbf{8.00} & \textbf{7.59}  &0.66 & 0.16 \\
 & WanS2V\cite{gao2025wans2vaudiodrivencinematicvideo} & 2.63 & 3.99 & 6.04 & 9.12  &0.80 & 0.25 \\
 & Ours & \textbf{3.42}& \textbf{4.76}& 7.16& 8.31& \textbf{0.97}& \textbf{45.2} \\
\bottomrule
\end{tabular}
}% end resizebox
\end{table}

\noindent
\textbf{Evaluation Metrics.}
We employ Q-Align~\cite{qalign} to evaluate perceptual quality (IQA) and aesthetic appeal (ASE). Audio-visual synchronization is measured via Sync-C and Sync-D~\cite{syn}. Identity consistency is assessed by DINOv2~\cite{oquab2023dinov2} cosine similarity (Dino-S) between generated frames and the reference image. For the in-domain AVSpeech evaluation (reported in the supplementary), we additionally include FID~\cite{fid} and FVD~\cite{fvd} since ground-truth videos are available.

\subsection{Comparison with Existing Methods}

We compare Live Avatar against current state-of-the-art open-sourced audio-driven avatar generation approaches, including Ditto~\cite{li2024ditto}, Echomimic-V2~\cite{meng2025echomimicv2strikingsimplifiedsemibody}, Hallo3~\cite{cui2025hallo3}, Stable-Avatar~\cite{tu2025stableavatar}, OmniAvatar~\cite{gan2025omniavatar}, and WanS2V~\cite{gao2025wans2vaudiodrivencinematicvideo}. All methods are benchmarked at 720$\times$400 on a single H100 node, with sequence-parallel methods using their maximum supported parallelism and all other settings kept at official defaults. Quantitative results on GenBench are presented in Table~\ref{tab:quantitative_comparison}.

\textbf{GenBench-ShortVideo.}
On short-form evaluation, our method achieves the best IQA and competitive ASE with OmniAvatar and Stable-Avatar, outperforming the rest. Although we build upon WanS2V and use step distillation, we surpass the teacher on visual quality; this aligns with the tendency of DMD-distilled models to concentrate the output distribution and attain slightly higher perceptual scores~\cite{imageteam2025zimage}, and the DMD loss has been interpreted as a specialized reinforcement learning objective that effectively optimizes both aesthetic appeal and foundational visual fidelity~\cite{luo2025learning,luo2024diff}. On Dino-S we obtain 0.96; Ditto's 0.99 is expected as it is a face-editing model that retains the reference face more than a full video-generation pipeline. Audio-visual synchronization is comparable to OmniAvatar (Sync-C and Sync-D are both strong for both methods). The most pronounced gap is inference speed: we reach 45.2 FPS. Ditto is the only other method with near real-time throughput (21.8 FPS); we use roughly 70$\times$ its parameters yet reach 45.2 FPS, which underscores the effectiveness of DMD, TPP, and our kernel-level optimizations.

\begin{figure}[t]
  \centering
   \includegraphics[width=0.98\linewidth]{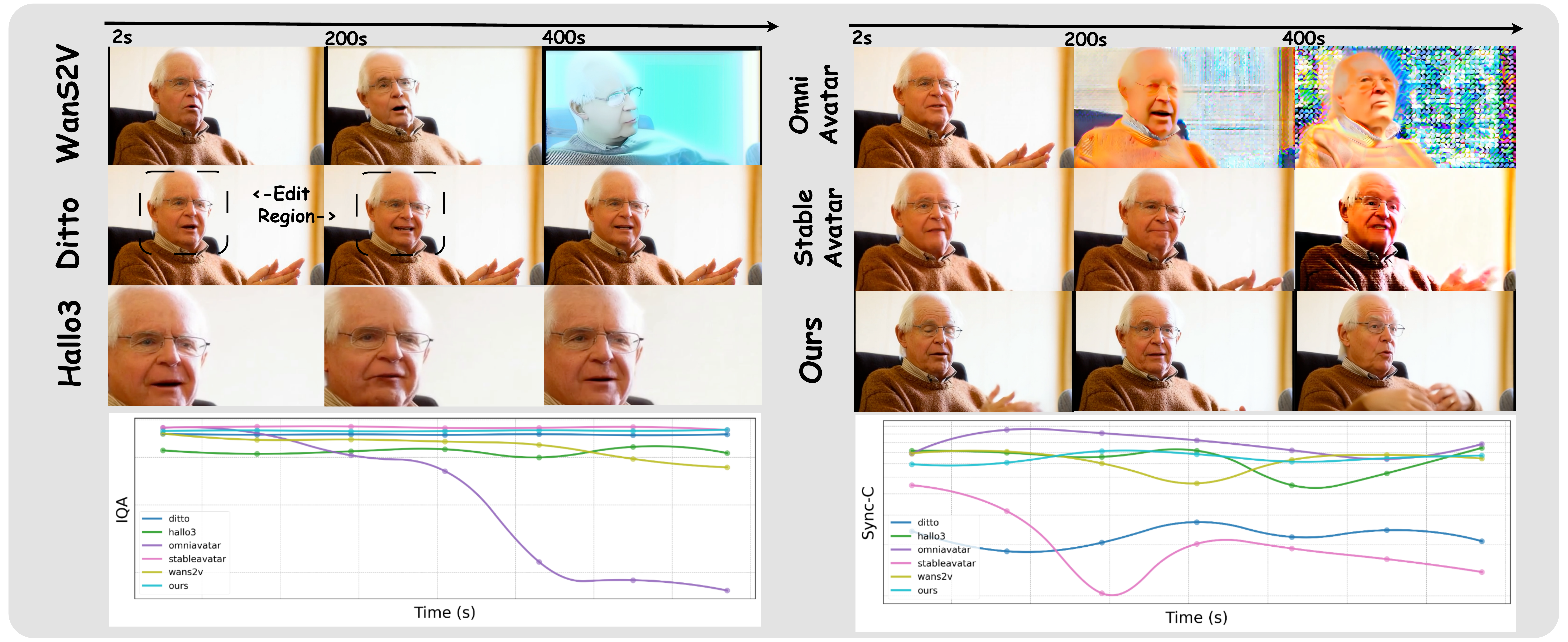}

   \caption{Qualitative comparisons with state-of-the-art methods.}
   \label{fig:Compare}
\end{figure}

\textbf{GenBench-LongVideo.}
Long-duration generation reveals a clear separation among methods. Echomimic-V2 is excluded as it does not support long-video inference. Comparing Short to Long results in Table~\ref{tab:quantitative_comparison}, most generative baselines degrade noticeably: OmniAvatar drops from 3.53/4.49 to 2.36/2.86 (ASE/IQA), WanS2V from 3.36/4.29 to 2.63/3.99, and Hallo3 from 3.12/3.97 to 2.65/4.04. Ditto, as a face-editing model, maintains relatively stable quality but lags in audio-visual synchronization (Sync-C 3.98). Our method, by contrast, preserves near-identical quality across durations (ASE 3.44$\rightarrow$3.42, IQA 4.51$\rightarrow$4.76, Dino-S 0.96$\rightarrow$0.97). OmniAvatar achieves the highest Sync-C/Sync-D on Long, but its severe visual collapse (Dino-S 0.66) suggests that loss of identity constraints artificially inflates sync scores. We achieve competitive synchronization (Sync-C 7.16) while maintaining the best overall visual and identity quality, yielding a more balanced profile across all metrics. Figure~\ref{fig:Compare} corroborates this: most baselines, especially OmniAvatar and WanS2V, exhibit visible quality degradation over time, whereas ours maintains stable fidelity throughout.

% 在这一节中，我们将详细分析Live Avatar中各种技术对于长视频生成效果和时延的影响。

% 我们分析Live Avatar模型架构和推理策略对于生成时间的影响，包括Timestep-forcing Pipelie Parallelism (TPP, in Sec. \ref{})、不用多卡并行策略（w/ SP4GPU）、不用VAE parallel策略、使用WanS2V的FramePack推理、只用Diffusion-forcing不用DMD。我们从两个维度分析推理效率：平均每秒视频帧生成时间FPS和 end-to-end time of generating first-frame (TTFF)，相关结果见Table xxx。可以看到，DMD由于NFE的大幅度减少，对于FPS和TTFF提升是首位的，不用TPP而是经典Self-forcing推理策略只会获得4FPS的效率，而如果用4卡的序列并行只会有轻微的效率提升，这是因为每个block的推理sequence较短（只有3个latent frames），attention等核心模块计算开销已经不是时延的主要因素。此外，FramePack的3个latent帧和VAE时延成为提升实时生成不可忽略的因素之一，故去掉FramePack且把VAE并行化策略成为实时视频生成的关键因素。
% % \begin{table*}[h]
% % \centering
% % \caption{Quantitative comparison on the test set [Placeholder].}
% % \label{tab:quantitative_comparison}
% % \begin{tabular}{lccccccl}
% % \toprule
% % \textbf{Methods} & \textbf{FID} $\downarrow$ & \textbf{FVD} $\downarrow$ & \textbf{Sync-C} $\uparrow$ & \textbf{Sync-D} $\downarrow$ & \textbf{IQA} $\uparrow$ & \textbf{ASE} $\uparrow$  &\textbf{FPS} $\uparrow$\\ 
% % \midrule
% % Hallo3 (Cui et al. 2024b) & 104 & 1078 & 5.23 & 9.54 & 3.41 & 2.00  &\\
% % FantasyTalking (Wang et al. 2025) & 78.9 & 780 & 3.14 & 11.2 & 3.33 & 1.96  &\\
% % HunyuanAvatar (Chen et al. 2025b) & 77.7 & 887 & 6.71 & 8.35 & 3.61 & 2.16  &\\
% % MultiTalk (Kong et al. 2025) & 74.7 & 787 & 4.76 & 9.99 & 3.67 & 2.22  &\\ 
% % \midrule
% % GT & - & - & 6.75 & 7.76 & 3.92 & 2.38  &\\
% % Ours & \textbf{67.6} & \textbf{664} & \textbf{7.12} & \textbf{8.05} & \textbf{3.75} & \textbf{2.25}  &20.8\\ 
% % \bottomrule
% % \end{tabular}
% % \end{table*}

\begin{table}[h]
\begin{minipage}[t]{0.52\columnwidth}
\centering
\captionof{table}{Ablation on Inference Efficiency. $SP_{4}$: sequence parallelism across 4 GPUs.}
\label{tab:Ablation_eff}
\resizebox{\linewidth}{!}{%
\setlength{\tabcolsep}{2pt}
\begin{tabular}{ccccc|cccc}
\toprule
DMD & $SP_{4}$ & TPP & VAE Para. & Kern. Opt. & \#GPU & NFE & FPS$\uparrow$ & TTFF$\downarrow$\\
\midrule
\XSolidBrush & \XSolidBrush & \XSolidBrush & \XSolidBrush & \XSolidBrush & 1 & 80 & 0.29 & 45.50\\
\CheckmarkBold & \XSolidBrush & \XSolidBrush & \XSolidBrush & \XSolidBrush & 1 & 5 & 3.66 & 4.56\\
\CheckmarkBold & \CheckmarkBold & \XSolidBrush & \XSolidBrush & \XSolidBrush & 4 & 5 & 4.50 & 3.94\\
\CheckmarkBold & \XSolidBrush & \CheckmarkBold & \XSolidBrush & \XSolidBrush & 4 & 4 & 10.16 & 4.73\\
\CheckmarkBold & \XSolidBrush & \CheckmarkBold & \CheckmarkBold & \XSolidBrush & 5 & 4 & 20.88 & 2.89\\
\midrule
\CheckmarkBold & \XSolidBrush & \CheckmarkBold & \CheckmarkBold & \CheckmarkBold & 5 & 4 & \textbf{45.2} & \textbf{1.21}\\
\bottomrule
\end{tabular}%
}
\end{minipage}
\hfill
\begin{minipage}[t]{0.47\columnwidth}
\centering
\captionof{table}{Ablation on Long Video Generation (GenBench-LongVideo).}
\label{tab:ablation_long}
\resizebox{\linewidth}{!}{%
\setlength{\tabcolsep}{3pt}
\begin{tabular}{l c c c c}
\toprule
\multicolumn{1}{c}{Methods} & \multicolumn{4}{c}{Metrics}\\
\cmidrule(lr){2-5}
 & ASE$\uparrow$ & IQA$\uparrow$ & Sync-C$\uparrow$ & Dino-S$\uparrow$\\
\midrule
w/o AAS & 3.13 & 4.68 & 7.25 & 0.96\\
w/o Rolling RoPE & 3.38 & \textbf{4.82} & \textbf{7.29} & 0.86\\
w/o History Corrupt & 2.90 & 3.88 & 7.14 & 0.81\\
Ours & \textbf{3.42} & 4.76 & 7.16 & \textbf{0.97}\\
\bottomrule
\end{tabular}%
}
\end{minipage}
\end{table}

\subsection{Ablation Study}
\label{sec:ablation}
\textbf{Study of inference efficiency.}
Table~\ref{tab:Ablation_eff} incrementally enables each component. DMD removes CFG and cuts NFE from 80 to 5 (4 denoising steps + 1 clean-cache pass needed by Self-Forcing~\cite{yin2025slow}). $SP_4$ provides only marginal speedup since per-block sequences are short (3 latents). TPP more than doubles throughput by pipelining steps across GPUs and eliminating the clean-cache pass (NFE 5$\to$4). VAE parallelization and kernel-level optimizations yield further gains, bringing the full system to 45.2 FPS / 1.21\,s TTFF on 5 H100 GPUs.

\textbf{Study on long-horizon stability strategies.}
As shown in Table~\ref{tab:ablation_long} and Figure~\ref{fig:rsfm_visual}, each component targets a distinct failure mode.
Removing History Corrupt causes the most severe degradation (ASE/IQA drop to 2.90/3.88, Dino-S to 0.81).
Without AAS, progressive color desaturation emerges, with noticeably grayed-out frames (ASE drops to 3.13; Figure~\ref{fig:rsfm_visual}).
Removing Rolling RoPE triggers identity drift (Dino-S 0.97$\to$0.86), with visible changes in hair details and facial features (Figure~\ref{fig:rsfm_visual}).

\textbf{Study on denoising step count.}
As shown in Table~\ref{tab:step_ablation}, increasing steps from 2 to 4 brings modest visual gains but markedly improves audio-visual synchronization (Sync-C 6.41$\to$7.03), since fewer steps leave insufficient budget for early, motion-critical denoising stages. Thanks to TPP, FPS stays at 45.2 regardless of step count; only TTFF grows (0.68\,s$\to$1.21\,s), which remains practical for interactive streaming where end-to-end latency is dominated by network transport. We therefore default to 4 steps.

\begin{table}[h]
\centering
\footnotesize
\caption{Effect of denoising step count on GenBench-ShortVideo. TPP keeps FPS constant regardless of step count; TTFF scales with the number of steps.}
\label{tab:step_ablation}
\begin{tabular}{l c c c c c c c}
\toprule
Method & ASE $\uparrow$ & IQA $\uparrow$ & Sync-C $\uparrow$ & Sync-D $\downarrow$ & Dino-S $\uparrow$ & TTFF (s) $\downarrow$ & FPS $\uparrow$ \\
\midrule
Ours (2 steps) & 3.37 & 4.25 & 6.41 & 9.02 & 0.94 & \textbf{0.68} & \textbf{45.2} \\
Ours (3 steps) & 3.41 & 4.36 & 6.58 & 8.85 & 0.95 & 0.95 & \textbf{45.2} \\
Ours (4 steps) & \textbf{3.44} & \textbf{4.51} & \textbf{7.03} & \textbf{8.30} & \textbf{0.96} & 1.21 & \textbf{45.2} \\
\bottomrule
\end{tabular}
\end{table}

\textbf{Study on noise level of KV cache.}
Table~\ref{tab:noise_level} compares three KV-cache noise schedules on GenBench-Long: clean-KV-cache, fix-noisy-KV-cache (noise fixed at $t\!=\!557$), and our timestep-forcing. Both baselines follow standard Self-Forcing inference~\cite{yin2025slow}. We additionally report T.Flicker~\cite{huang2023vbench}, which measures temporal consistency. Clean-KV-cache degrades rapidly over long horizons due to error accumulation. Fixed noise improves quality by decoupling identity and motion dynamics in the cache~\cite{song2025historyguidedvideodiffusion}, but both baselines suffer from severe flickering (T.Flicker 0.876/0.891). Only timestep-forcing resolves this (T.Flicker 0.971) by maintaining step-matched caches. Furthermore, timestep-forcing is the only schedule compatible with TPP.

\begin{table}[t]
\centering
\footnotesize
\caption{Effect of KV-cache noise-level scheduling on GenBench-Long.}
\label{tab:noise_level}
\setlength{\tabcolsep}{3pt}
\resizebox{\columnwidth}{!}{%
\begin{tabular}{l c c c c c c}
\toprule
\multicolumn{1}{c}{Setting} & \multicolumn{6}{c}{Metrics} \\
\cmidrule(lr){2-7}
 & ASE $\uparrow$ & IQA $\uparrow$ & Sync-C $\uparrow$ & Sync-D $\downarrow$ & Dino-S $\uparrow$ & T.Flicker$\uparrow$ \\
\midrule
clean-kv-cache & 3.05 & 4.44 & 6.11 & 9.10 & 0.90 & 0.876 \\
fix-noisy-kv-cache & 3.34 & 4.63 & 7.00 & 8.38 & 0.93 & 0.891 \\
timestep-forcing & \textbf{3.42} & \textbf{4.76} & \textbf{7.16} & \textbf{8.31} & \textbf{0.97} & \textbf{0.971} \\
\bottomrule
\end{tabular}%
}
\end{table}

\begin{figure}[t]
  \centering
  \includegraphics[width=\linewidth]{images/main/rsfm_ablation_visual.pdf}
  \caption{Visual ablation of long-horizon stability components on long video generation.}
  \label{fig:rsfm_visual}
\end{figure}

\noindent
\begin{minipage}[t]{0.48\linewidth}
\textbf{Study on motion-frame scaffolding.}
Starting from the same 25\,K-step Stage~1 with all other hyperparameters aligned, we compare Stage~2 convergence with and without motion-frame scaffolding (Figure~\ref{fig:scaffold_conv}). The scaffolded variant saturates at $\sim$500 steps, while the baseline requires $\sim$2500 steps---a $5\times$ speedup with no loss in final quality.
\end{minipage}%
\hfill
\begin{minipage}[t]{0.5\linewidth}
\strut\par\centering
\includegraphics[width=0.8\linewidth]{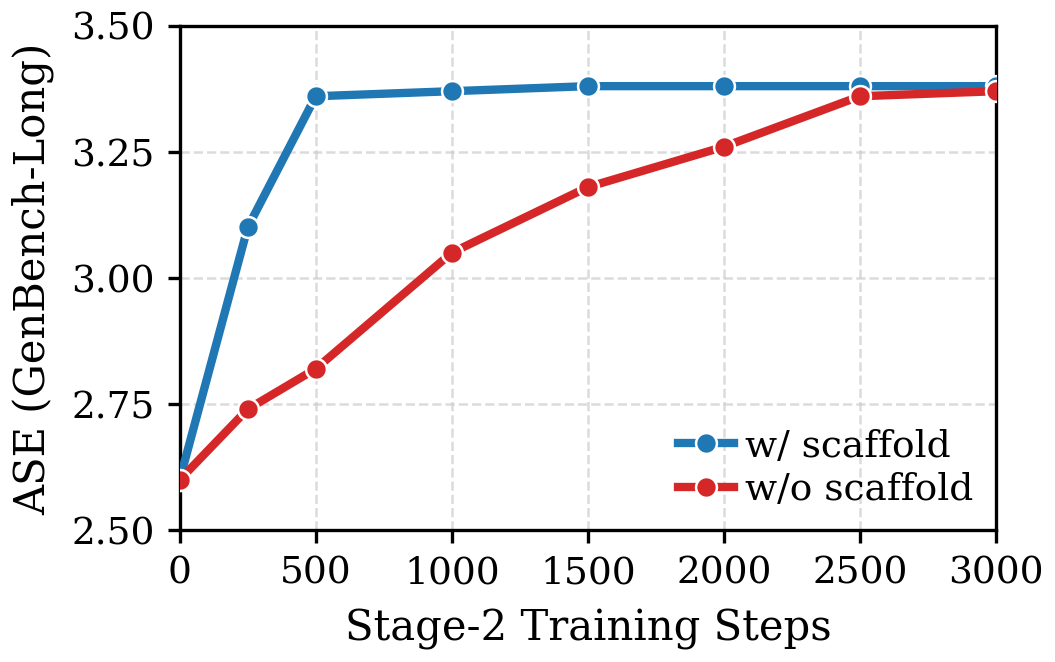}
\captionof{figure}{Study of motion-frame scaffolding.}
\label{fig:scaffold_conv}
\end{minipage}

% ablation table1:不同id策略的长视频曲线，长视频质量 （-更新参考帧，-rolling sink frame, 参考帧距离？，kv_cache加噪
% ablation table2:
% 不同推理策略的fps和tff
% -timestep paral , 
% -timestep paral + sp
% + sp
% -vae paral
% ours

\section{Conclusion}
We present \textbf{Live Avatar}, an algorithm-system co-designed framework that, to our knowledge, is the first to enable practical real-time streaming of a 14-billion-parameter diffusion model for infinite-length audio-driven avatar generation. On the algorithm side, a two-stage training pipeline adapts a pretrained bidirectional model into a causal, few-step streaming one. Three complementary long-horizon strategies---History Corrupt, Adaptive Attention Sink, and Rolling RoPE---enable stable generation beyond 10{,}000 seconds\footnote{We provide experimental results on 10{,}000-second generation in the supplementary material.}. On the system side, Timestep-forcing Pipeline Parallelism (TPP) assigns each GPU a fixed denoising timestep, converting the sequential diffusion chain into an asynchronous spatial pipeline that achieves 45 FPS with a TTFF of 1.21\,s on 5 H100 GPUs. We also introduce GenBench, a long-form benchmark exceeding five minutes, to support reproducible evaluation.

% \noindent\textbf{Limitation.}
% TPP scales throughput linearly with denoising steps but does not reduce TTFF, limiting first-response latency in interactive settings.
\noindent\textbf{Limitation.}
Our long-horizon stability strategies are grounded in the static-scene prior inherent to avatar interaction, where the subject and background remain largely unchanged; as native long-duration training becomes feasible, such priors may become unnecessary. Additionally, the current TTFF of 1.21\,s, combined with network transport overhead, yields an end-to-end latency of approximately 3\,s, which does not yet meet the stringent requirements of seamless bidirectional interaction.
% % %limitation:
% % %一是并行加速提帧率不提延时 
% % %二是过度依赖参考帧出现很多时序不一致的瑕疵，且损害模型的长时序记忆能力，

\bibliographystyle{splncs04}
\bibliography{6_references}

@inproceedings{yin2025slow,
  title={From slow bidirectional to fast autoregressive video diffusion models},
  author={Yin, Tianwei and Zhang, Qiang and Zhang, Richard and Freeman, William T and Durand, Fredo and Shechtman, Eli and Huang, Xun},
  booktitle={Proceedings of the Computer Vision and Pattern Recognition Conference},
  pages={22963--22974},
  year={2025}
}

@article{chen2024diffusion,
  title={Diffusion forcing: Next-token prediction meets full-sequence diffusion},
  author={Chen, Boyuan and Mart{\'\i} Mons{\'o}, Diego and Du, Yilun and Simchowitz, Max and Tedrake, Russ and Sitzmann, Vincent},
  journal={Advances in Neural Information Processing Systems},
  volume={37},
  pages={24081--24125},
  year={2024}
}

@article{huang2025self,
  title={Self Forcing: Bridging the Train-Test Gap in Autoregressive Video Diffusion},
  author={Huang, Xun and Li, Zhengqi and He, Guande and Zhou, Mingyuan and Shechtman, Eli},
  journal={arXiv preprint arXiv:2506.08009},
  year={2025}
}

@article{cui2025self,
  title={Self-Forcing++: Towards Minute-Scale High-Quality Video Generation},
  author={Cui, Justin and Wu, Jie and Li, Ming and Yang, Tao and Li, Xiaojie and Wang, Rui and Bai, Andrew and Ban, Yuanhao and Hsieh, Cho-Jui},
  journal={arXiv preprint arXiv:2510.02283},
  year={2025}
}

@article{yang2025longlive,
  title={Longlive: Real-time interactive long video generation},
  author={Yang, Shuai and Huang, Wei and Chu, Ruihang and Xiao, Yicheng and Zhao, Yuyang and Wang, Xianbang and Li, Muyang and Xie, Enze and Chen, Yingcong and Lu, Yao and others},
  journal={arXiv preprint arXiv:2509.22622},
  year={2025}
}

@article{liu2025rolling,
  title={Rolling Forcing: Autoregressive Long Video Diffusion in Real Time},
  author={Liu, Kunhao and Hu, Wenbo and Xu, Jiale and Shan, Ying and Lu, Shijian},
  journal={arXiv preprint arXiv:2509.25161},
  year={2025}
}

@inproceedings{prajwal2020lip,
  title={A lip sync expert is all you need for speech to lip generation in the wild},
  author={Prajwal, KR and Mukhopadhyay, Rudrabha and Namboodiri, Vinay P and Jawahar, CV},
  booktitle={Proceedings of the 28th ACM international conference on multimedia},
  pages={484--492},
  year={2020}
}

@inproceedings{tian2024emo,
  title={Emo: Emote portrait alive generating expressive portrait videos with audio2video diffusion model under weak conditions},
  author={Tian, Linrui and Wang, Qi and Zhang, Bang and Bo, Liefeng},
  booktitle={European Conference on Computer Vision},
  pages={244--260},
  year={2024},
  organization={Springer}
}

@article{yu2025llia,
  title={LLIA--Enabling Low-Latency Interactive Avatars: Real-Time Audio-Driven Portrait Video Generation with Diffusion Models},
  author={Yu, Haojie and Wang, Zhaonian and Pan, Yihan and Cheng, Meng and Yang, Hao and Wang, Chao and Xie, Tao and Xu, Xiaoming and Wei, Xiaoming and Cai, Xunliang},
  journal={arXiv preprint arXiv:2506.05806},
  year={2025}
}

@inproceedings{zhang2023sadtalker,
  title={Sadtalker: Learning realistic 3d motion coefficients for stylized audio-driven single image talking face animation},
  author={Zhang, Wenxuan and Cun, Xiaodong and Wang, Xuan and Zhang, Yong and Shen, Xi and Guo, Yu and Shan, Ying and Wang, Fei},
  booktitle={Proceedings of the IEEE/CVF conference on computer vision and pattern recognition},
  pages={8652--8661},
  year={2023}
}

@article{videoworldsimulators2024,
  title={Video generation models as world simulators},
  author={Tim Brooks and Bill Peebles and Connor Holmes and Will DePue and Yufei Guo and Li Jing and David Schnurr and Joe Taylor and Troy Luhman and Eric Luhman and Clarence Ng and Ricky Wang and Aditya Ramesh},
  year={2024},
  url={https://openai.com/research/video-generation-models-as-world-simulators},
}

@article{hacohen2024ltx,
  title={Ltx-video: Realtime video latent diffusion},
  author={HaCohen, Yoav and Chiprut, Nisan and Brazowski, Benny and Shalem, Daniel and Moshe, Dudu and Richardson, Eitan and Levin, Eran and Shiran, Guy and Zabari, Nir and Gordon, Ori and others},
  journal={arXiv preprint arXiv:2501.00103},
  year={2024}
}

@article{kong2024hunyuanvideo,
  title={Hunyuanvideo: A systematic framework for large video generative models},
  author={Kong, Weijie and Tian, Qi and Zhang, Zijian and Min, Rox and Dai, Zuozhuo and Zhou, Jin and Xiong, Jiangfeng and Li, Xin and Wu, Bo and Zhang, Jianwei and others},
  journal={arXiv preprint arXiv:2412.03603},
  year={2024}
}

@article{wan2025wan,
  title={Wan: Open and advanced large-scale video generative models},
  author={Wan, Team and Wang, Ang and Ai, Baole and Wen, Bin and Mao, Chaojie and Xie, Chen-Wei and Chen, Di and Yu, Feiwu and Zhao, Haiming and Yang, Jianxiao and others},
  journal={arXiv preprint arXiv:2503.20314},
  year={2025}
}

@article{du2025rap,
  title={RAP: Real-time Audio-driven Portrait Animation with Video Diffusion Transformer},
  author={Du, Fangyu and Li, Taiqing and Zhang, Ziwei and Qiao, Qian and Yu, Tan and Zhen, Dingcheng and Jia, Xu and Yang, Yang and Yin, Shunshun and Liu, Siyuan},
  journal={arXiv preprint arXiv:2508.05115},
  year={2025}
}

@article{meng2025mirrorme,
  title={MirrorMe: Towards Realtime and High Fidelity Audio-Driven Halfbody Animation},
  author={Meng, Dechao and Xiao, Steven and Zhang, Xindi and Wang, Guangyuan and Zhang, Peng and Wang, Qi and Zhang, Bang and Bo, Liefeng},
  journal={arXiv preprint arXiv:2506.22065},
  year={2025}
}

@article{li2024ditto,
  title={Ditto: Motion-space diffusion for controllable realtime talking head synthesis},
  author={Li, Tianqi and Zheng, Ruobing and Yang, Minghui and Chen, Jingdong and Yang, Ming},
  journal={arXiv preprint arXiv:2411.19509},
  year={2024}
}

@article{wang2025omnitalker,
  title={Omnitalker: Real-time text-driven talking head generation with in-context audio-visual style replication},
  author={Wang, Zhongjian and Zhang, Peng and Qi, Jinwei and Xu, Guangyuan Wang Sheng and Zhang, Bang and Bo, Liefeng},
  journal={arXiv e-prints},
  pages={arXiv--2504},
  year={2025}
}

@inproceedings{zhen2025teller,
  title={Teller: Real-Time Streaming Audio-Driven Portrait Animation with Autoregressive Motion Generation},
  author={Zhen, Dingcheng and Yin, Shunshun and Qin, Shiyang and Yi, Hou and Zhang, Ziwei and Liu, Siyuan and Qi, Gan and Tao, Ming},
  booktitle={Proceedings of the Computer Vision and Pattern Recognition Conference},
  pages={21075--21085},
  year={2025}
}

@article{guo2025arig,
  title={ARIG: Autoregressive Interactive Head Generation for Real-time Conversations},
  author={Guo, Ying and Liu, Xi and Zhen, Cheng and Yan, Pengfei and Wei, Xiaoming},
  journal={arXiv preprint arXiv:2507.00472},
  year={2025}
}

@inproceedings{zhu2025infp,
  title={INFP: Audio-driven interactive head generation in dyadic conversations},
  author={Zhu, Yongming and Zhang, Longhao and Rong, Zhengkun and Hu, Tianshu and Liang, Shuang and Ge, Zhipeng},
  booktitle={Proceedings of the Computer Vision and Pattern Recognition Conference},
  pages={10667--10677},
  year={2025}
}

@article{kodaira2025streamdit,
  title={Streamdit: Real-time streaming text-to-video generation},
  author={Kodaira, Akio and Hou, Tingbo and Hou, Ji and Tomizuka, Masayoshi and Zhao, Yue},
  journal={arXiv preprint arXiv:2507.03745},
  year={2025}
}

@article{li2024autoregressive,
  title={Autoregressive image generation without vector quantization},
  author={Li, Tianhong and Tian, Yonglong and Li, He and Deng, Mingyang and He, Kaiming},
  journal={Advances in Neural Information Processing Systems},
  volume={37},
  pages={56424--56445},
  year={2024}
}

@article{ao2024body,
  title={Body of her: A preliminary study on end-to-end humanoid agent},
  author={Ao, Tenglong},
  journal={arXiv preprint arXiv:2408.02879},
  year={2024}
}

@article{chen2025midas,
  title={Midas: Multimodal interactive digital-human synthesis via real-time autoregressive video generation},
  author={Chen, Ming and Cui, Liyuan and Zhang, Wenyuan and Zhang, Haoxian and Zhou, Yan and Li, Xiaohan and Tang, Songlin and Liu, Jiwen and Liao, Borui and Chen, Hejia and others},
  journal={arXiv preprint arXiv:2508.19320},
  year={2025}
}

@article{xie2025x,
  title={X-Streamer: Unified Human World Modeling with Audiovisual Interaction},
  author={Xie, You and Gu, Tianpei and Li, Zenan and Zhang, Chenxu and Song, Guoxian and Zhao, Xiaochen and Liang, Chao and Jiang, Jianwen and Xu, Hongyi and Luo, Linjie},
  journal={arXiv preprint arXiv:2509.21574},
  year={2025}
}

@article{frans2024one,
  title={One step diffusion via shortcut models},
  author={Frans, Kevin and Hafner, Danijar and Levine, Sergey and Abbeel, Pieter},
  journal={arXiv preprint arXiv:2410.12557},
  year={2024}
}

@inproceedings{yin2024one,
  title={One-step diffusion with distribution matching distillation},
  author={Yin, Tianwei and Gharbi, Micha{\"e}l and Zhang, Richard and Shechtman, Eli and Durand, Fredo and Freeman, William T and Park, Taesung},
  booktitle={Proceedings of the IEEE/CVF conference on computer vision and pattern recognition},
  pages={6613--6623},
  year={2024}
}

@article{song2023consistency,
  title={Consistency models},
  author={Song, Yang and Dhariwal, Prafulla and Chen, Mark and Sutskever, Ilya},
  year={2023}
}

@article{luo2025learning,
  title={Learning Few-Step Diffusion Models by Trajectory Distribution Matching},
  author={Luo, Yihong and Hu, Tianyang and Sun, Jiacheng and Cai, Yujun and Tang, Jing},
  journal={arXiv preprint arXiv:2503.06674},
  year={2025}
}

@article{lu2024simplifying,
  title={Simplifying, stabilizing and scaling continuous-time consistency models},
  author={Lu, Cheng and Song, Yang},
  journal={arXiv preprint arXiv:2410.11081},
  year={2024}
}

@article{zheng2025large,
  title={Large Scale Diffusion Distillation via Score-Regularized Continuous-Time Consistency},
  author={Zheng, Kaiwen and Wang, Yuji and Ma, Qianli and Chen, Huayu and Zhang, Jintao and Balaji, Yogesh and Chen, Jianfei and Liu, Ming-Yu and Zhu, Jun and Zhang, Qinsheng},
  journal={arXiv preprint arXiv:2510.08431},
  year={2025}
}

@article{luo2024diff,
  title={Diff-instruct++: Training one-step text-to-image generator model to align with human preferences},
  author={Luo, Weijian},
  journal={arXiv preprint arXiv:2410.18881},
  year={2024}
}

@article{luo2023diff,
  title={Diff-instruct: A universal approach for transferring knowledge from pre-trained diffusion models},
  author={Luo, Weijian and Hu, Tianyang and Zhang, Shifeng and Sun, Jiacheng and Li, Zhenguo and Zhang, Zhihua},
  journal={Advances in Neural Information Processing Systems},
  volume={36},
  pages={76525--76546},
  year={2023}
}

@article{yin2024improved,
  title={Improved distribution matching distillation for fast image synthesis},
  author={Yin, Tianwei and Gharbi, Micha{\"e}l and Park, Taesung and Zhang, Richard and Shechtman, Eli and Durand, Fredo and Freeman, Bill},
  journal={Advances in neural information processing systems},
  volume={37},
  pages={47455--47487},
  year={2024}
}

@article{lipman2022flow,
  title={Flow matching for generative modeling},
  author={Lipman, Yaron and Chen, Ricky TQ and Ben-Hamu, Heli and Nickel, Maximilian and Le, Matt},
  journal={arXiv preprint arXiv:2210.02747},
  year={2022}
}

@article{gan2025omniavatar,
  title={OmniAvatar: Efficient Audio-Driven Avatar Video Generation with Adaptive Body Animation},
  author={Gan, Qijun and Yang, Ruizi and Zhu, Jianke and Xue, Shaofei and Hoi, Steven},
  journal={arXiv preprint arXiv:2506.18866},
  year={2025}
}

@inproceedings{cui2025hallo3,
  title={Hallo3: Highly dynamic and realistic portrait image animation with video diffusion transformer},
  author={Cui, Jiahao and Li, Hui and Zhan, Yun and Shang, Hanlin and Cheng, Kaihui and Ma, Yuqi and Mu, Shan and Zhou, Hang and Wang, Jingdong and Zhu, Siyu},
  booktitle={Proceedings of the Computer Vision and Pattern Recognition Conference},
  pages={21086--21095},
  year={2025}
}

@inproceedings{wang2025fantasytalking,
  title={Fantasytalking: Realistic talking portrait generation via coherent motion synthesis},
  author={Wang, Mengchao and Wang, Qiang and Jiang, Fan and Fan, Yaqi and Zhang, Yunpeng and Qi, Yonggang and Zhao, Kun and Xu, Mu},
  booktitle={Proceedings of the 33rd ACM International Conference on Multimedia},
  pages={9891--9900},
  year={2025}
}

@article{fid,
  title={Gans trained by a two time-scale update rule converge to a local nash equilibrium},
  author={Heusel, Martin and Ramsauer, Hubert and Unterthiner, Thomas and Nessler, Bernhard and Hochreiter, Sepp},
  journal={Advances in neural information processing systems},
  volume={30},
  year={2017}
}

@article{fvd,
  title={Towards accurate generative models of video: A new metric \& challenges},
  author={Unterthiner, Thomas and Van Steenkiste, Sjoerd and Kurach, Karol and Marinier, Raphael and Michalski, Marcin and Gelly, Sylvain},
  journal={arXiv preprint arXiv:1812.01717},
  year={2018}
}

@article{qalign,
  title={Q-align: Teaching lmms for visual scoring via discrete text-defined levels},
  author={Wu, Haoning and Zhang, Zicheng and Zhang, Weixia and Chen, Chaofeng and Liao, Liang and Li, Chunyi and Gao, Yixuan and Wang, Annan and Zhang, Erli and Sun, Wenxiu and others},
  journal={arXiv preprint arXiv:2312.17090},
  year={2023}
}

@inproceedings{syn,
  title={Out of time: automated lip sync in the wild},
  author={Chung, Joon Son and Zisserman, Andrew},
  booktitle={Computer Vision--ACCV 2016 Workshops: ACCV 2016 International Workshops, Taipei, Taiwan, November 20-24, 2016, Revised Selected Papers, Part II 13},
  pages={251--263},
  year={2017},
  organization={Springer}
}

@article{AVSpeech,
  title={Looking to listen at the cocktail party: A speaker-independent audio-visual model for speech separation},
  author={Ephrat, Ariel and Mosseri, Inbar and Lang, Oran and Dekel, Tali and Wilson, Kevin and Hassidim, Avinatan and Freeman, William T and Rubinstein, Michael},
  journal={arXiv preprint arXiv:1804.03619},
  year={2018}
}

@article{oquab2023dinov2,
  title={DINOv2: Learning Robust Visual Features without Supervision},
  author={Oquab, Maxime and Darcet, Timoth{\'e}e and Moutakanni, Th{\'e}o and Vo, Huy V. and Szafraniec, Marc and Khalidov, Vasil and Fernandez, Pierre and Haziza, Daniel and Massa, Francisco and El-Nouby, Alaaeldin and Assran, Mahmoud and Ballas, Nicolas and Galuba, Wojciech and Howes, Russell and Huang, Po-Yao and Li, Shang-Wen and Misra, Ishan and Rabbat, Michael and Sharma, Vasu and Synnaeve, Gabriel and Xu, Hu and J{\'e}gou, Herv{\'e} and Mairal, Julien and Labatut, Patrick and Joulin, Armand and Bojanowski, Piotr},
  journal={Transactions on Machine Learning Research},
  year={2024},
  url={https://arxiv.org/abs/2304.07193}
}

@misc{wu2025qwenimagetechnicalreport,
      title={Qwen-Image Technical Report}, 
      author={Chenfei Wu and Jiahao Li and Jingren Zhou and Junyang Lin and Kaiyuan Gao and Kun Yan and Sheng-ming Yin and Shuai Bai and Xiao Xu and Yilei Chen and Yuxiang Chen and Zecheng Tang and Zekai Zhang and Zhengyi Wang and An Yang and Bowen Yu and Chen Cheng and Dayiheng Liu and Deqing Li and Hang Zhang and Hao Meng and Hu Wei and Jingyuan Ni and Kai Chen and Kuan Cao and Liang Peng and Lin Qu and Minggang Wu and Peng Wang and Shuting Yu and Tingkun Wen and Wensen Feng and Xiaoxiao Xu and Yi Wang and Yichang Zhang and Yongqiang Zhu and Yujia Wu and Yuxuan Cai and Zenan Liu},
      year={2025},
      eprint={2508.02324},
      archivePrefix={arXiv},
      primaryClass={cs.CV},
      url={https://arxiv.org/abs/2508.02324}, 
}

@misc{du2024cosyvoice2scalablestreaming,
      title={CosyVoice 2: Scalable Streaming Speech Synthesis with Large Language Models}, 
      author={Zhihao Du and Yuxuan Wang and Qian Chen and Xian Shi and Xiang Lv and Tianyu Zhao and Zhifu Gao and Yexin Yang and Changfeng Gao and Hui Wang and Fan Yu and Huadai Liu and Zhengyan Sheng and Yue Gu and Chong Deng and Wen Wang and Shiliang Zhang and Zhijie Yan and Jingren Zhou},
      year={2024},
      eprint={2412.10117},
      archivePrefix={arXiv},
      primaryClass={cs.SD},
      url={https://arxiv.org/abs/2412.10117}, 
}

@misc{quignon2025thevalevaluationframeworktalking,
      title={THEval. Evaluation Framework for Talking Head Video Generation}, 
      author={Nabyl Quignon and Baptiste Chopin and Yaohui Wang and Antitza Dantcheva},
      year={2025},
      eprint={2511.04520},
      archivePrefix={arXiv},
      primaryClass={cs.CV},
      url={https://arxiv.org/abs/2511.04520}, 
}

@article{low2025talkingmachines,
  title={TalkingMachines: Real-Time Audio-Driven FaceTime-Style Video via Autoregressive Diffusion Models},
  author={Low, Chetwin and Wang, Weimin},
  journal={arXiv preprint arXiv:2506.03099},
  year={2025}
}

@article{tu2025stableavatar,
  title={Stableavatar: Infinite-length audio-driven avatar video generation},
  author={Tu, Shuyuan and Pan, Yueming and Huang, Yinming and Han, Xintong and Xing, Zhen and Dai, Qi and Luo, Chong and Wu, Zuxuan and Jiang, Yu-Gang},
  journal={arXiv preprint arXiv:2508.08248},
  year={2025}
}

@article{yang2025infinitetalk,
  title={InfiniteTalk: Audio-driven Video Generation for Sparse-Frame Video Dubbing},
  author={Yang, Shaoshu and Kong, Zhe and Gao, Feng and Cheng, Meng and Liu, Xiangyu and Zhang, Yong and Kang, Zhuoliang and Luo, Wenhan and Cai, Xunliang and He, Ran and others},
  journal={arXiv preprint arXiv:2508.14033},
  year={2025}
}

@misc{meng2025echomimicv2strikingsimplifiedsemibody,
      title={EchoMimicV2: Towards Striking, Simplified, and Semi-Body Human Animation}, 
      author={Rang Meng and Xingyu Zhang and Yuming Li and Chenguang Ma},
      year={2025},
      eprint={2411.10061},
      archivePrefix={arXiv},
      primaryClass={cs.GR},
      url={https://arxiv.org/abs/2411.10061}, 
}

@misc{gao2025wans2vaudiodrivencinematicvideo,
      title={Wan-S2V: Audio-Driven Cinematic Video Generation}, 
      author={Xin Gao and Li Hu and Siqi Hu and Mingyang Huang and Chaonan Ji and Dechao Meng and Jinwei Qi and Penchong Qiao and Zhen Shen and Yafei Song and Ke Sun and Linrui Tian and Guangyuan Wang and Qi Wang and Zhongjian Wang and Jiayu Xiao and Sheng Xu and Bang Zhang and Peng Zhang and Xindi Zhang and Zhe Zhang and Jingren Zhou and Lian Zhuo},
      year={2025},
      eprint={2508.18621},
      archivePrefix={arXiv},
      primaryClass={cs.CV},
      url={https://arxiv.org/abs/2508.18621}, 
}

@misc{song2025historyguidedvideodiffusion,
  title={History-Guided Video Diffusion},
  author={Kiwhan Song and Boyuan Chen and Max Simchowitz and Yilun Du and Russ Tedrake and Vincent Sitzmann},
  year={2025},
  eprint={2502.06764},
  archivePrefix={arXiv},
  primaryClass={cs.LG},
  url={https://arxiv.org/abs/2502.06764}
}

@article{song2019generative,
  author       = {Yang Song and Stefano Ermon},
  title        = {Generative Modeling by Estimating Gradients of the Data Distribution},
  journal      = {CoRR},
  volume       = {abs/1907.05600},
  year         = {2019},
  url          = {http://arxiv.org/abs/1907.05600},
  eprinttype   = {arXiv},
  eprint       = {1907.05600}
}

@misc{gelberg2025extending,
  title={Extending the Context of Pretrained LLMs by Dropping Their Positional Embeddings},
  author={Yoav Gelberg and Koshi Eguchi and Takuya Akiba and Edoardo Cetin},
  year={2025},
  eprint={2512.12167},
  archivePrefix={arXiv},
  primaryClass={cs.CL},
  url={https://arxiv.org/abs/2512.12167}
}

@misc{huang2023vbench,
  title={VBench: Comprehensive Benchmark Suite for Video Generative Models},
  author={Ziqi Huang and Yinan He and Jiashuo Yu and Fan Zhang and Chenyang Si and Yuming Jiang and Yuanhan Zhang and Tianxing Wu and Qingyang Jin and Nattapol Chanpaisit and Yaohui Wang and Xinyuan Chen and Limin Wang and Dahua Lin and Yu Qiao and Ziwei Liu},
  year={2023},
  eprint={2311.17982},
  archivePrefix={arXiv},
  primaryClass={cs.CV},
  url={https://arxiv.org/abs/2311.17982}
}

@misc{imageteam2025zimage,
      title={Z-Image: An Efficient Image Generation Foundation Model with Single-Stream Diffusion Transformer}, 
      author={Image Team and Huanqia Cai and Sihan Cao and Ruoyi Du and Peng Gao and Steven Hoi and Zhaohui Hou and Shijie Huang and Dengyang Jiang and Xin Jin and Liangchen Li and Zhen Li and Zhong-Yu Li and David Liu and Dongyang Liu and Junhan Shi and Qilong Wu and Feng Yu and Chi Zhang and Shifeng Zhang and Shilin Zhou},
      year={2025},
      eprint={2511.22699},
      archivePrefix={arXiv},
      primaryClass={cs.CV},
      url={https://arxiv.org/abs/2511.22699}, 
}

\clearpage
\appendix

\section{Overview of Supplementary Material}
\label{sec:overview}

\begin{figure*}[t!]
  \centering
   \includegraphics[width=0.98\linewidth]{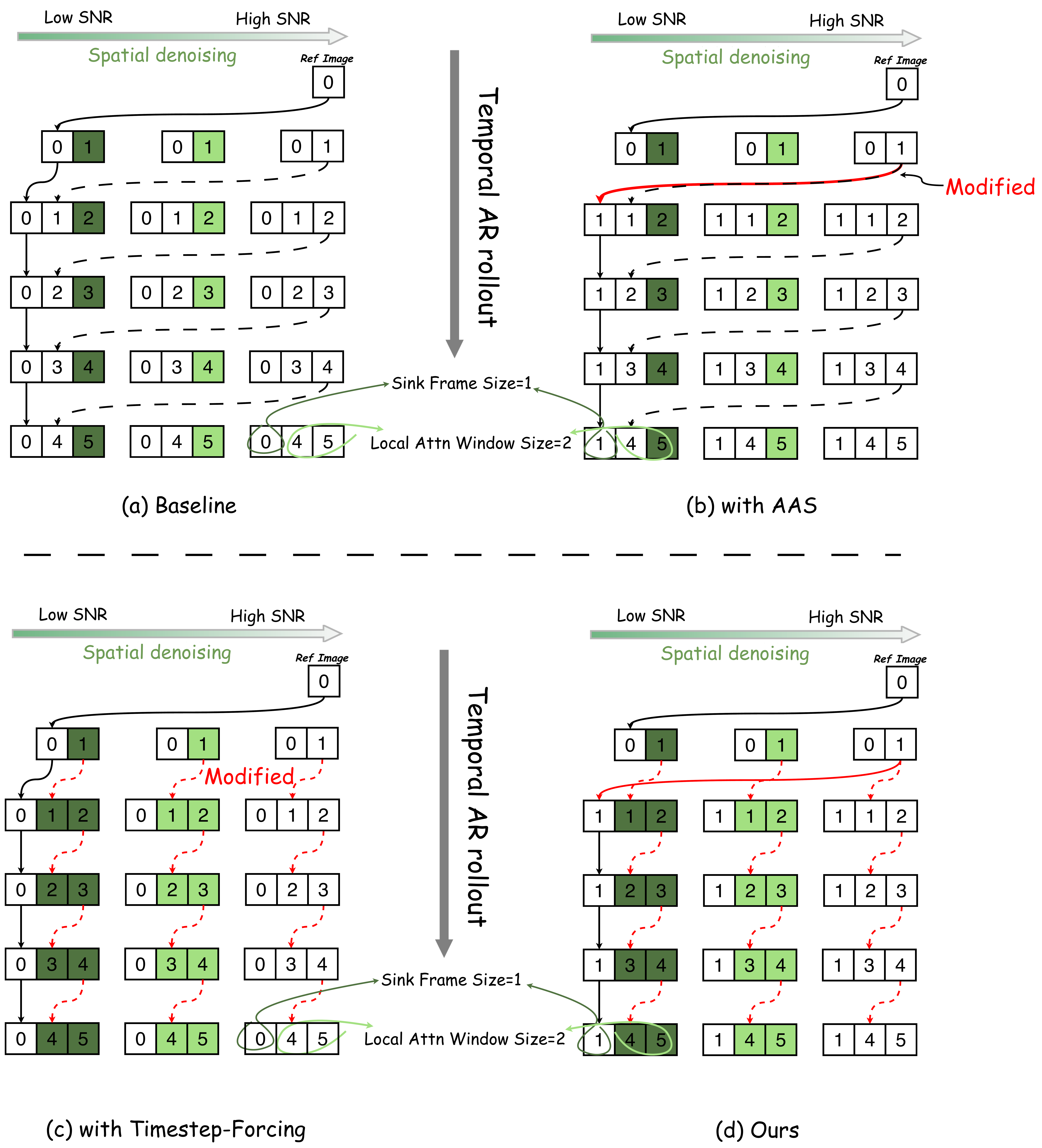}
   \caption{Illustration of different inference settings. Horizontally, each row follows the spatial denoising order from low to high SNR; vertically, each column shows the autoregressive rollout over frames. Each small \textbf{rectangle} denotes the latent of a block, and the \textbf{number} inside represents its block index. \textbf{Solid arrows} indicate direct sink frame passing, whereas \textbf{dashed arrows} indicate KV-cache passing. \textbf{Red} marks indicate the components modified relative to the baseline. (a) Baseline with a fixed sink frame and standard rolling-kv-cache. (b) \textbf{AAS} with the sink replaced by the first generated latent. (c) \textbf{Timestep-forcing} with each noisy latent attending only to KV caches of the same timestep. (d) \textbf{Ours} with both AAS and timestep-forcing.
   }
   \label{fig:method3}
\end{figure*}

This supplementary document provides comprehensive details, additional experiments, and implementation specifics to support the findings in the main paper. The content is organized as follows:

\begin{itemize}
\item \textbf{Section \ref{sec:additional_exp}: Additional Experimental Results.} We evaluate our model's long-horizon autoregressive extrapolation up to \textit{10,000}\,s (Table~\ref{tab:supp3}, Figure~\ref{fig:visual1}), provide additional comparison against state-of-the-art on the AV-Speech test set (Table~\ref{tab:supp2}), present a user study (Table~\ref{tab:user_study}, Figure~\ref{fig:user-study}), and ablate the effect of block size (Table~\ref{tab:block_size}).

\item \textbf{Section \ref{sec:additional setting}: Additional Evaluation Details.} We detail the hardware and inference configurations and runtime metric definitions used in our benchmarking.

\item \textbf{Section \ref{sec:kernel_opt}: Kernel-Level Optimizations.} We describe the kernel-level optimizations (referred to as \textit{Kernel Opt.}\ in the efficiency ablation of the main paper) that enable real-time streaming inference, including FP8 quantization, FlashAttention-3, cuDNN fused attention, torch.compile, LoRA weight merging, streaming VAE decoding, and model offloading.

\item \textbf{Section \ref{sec:algo}: Algorithm Details.} We provide complete pseudocode (Algorithm~\ref{alg:self_forcing1},~\ref{alg:self_forcing2},~\ref{alg:inference1}, and~\ref{alg:inference2}) for our methods, including the \textit{History Corrupt} and \textit{Block-wise Gradient Accumulation} training strategies. We also provide detailed implementation descriptions and inference procedures for \textit{AAS} and \textit{Rolling RoPE}, as well as \textit{TPP}. We further include visualizations (Figures~\ref{fig:method3},~\ref{fig:method4}, and~\ref{fig:TPP}) to illustrate the mechanisms of different inference settings.

\item \textbf{Section \ref{sec:visuals}: Additional Visual Results.} We showcase further qualitative examples to demonstrate the temporal consistency and visual fidelity of our method.

\item \textbf{Section \ref{sec:ethical}: Ethical Consideration.} We discuss the potential societal impacts, privacy concerns, and responsible usage guidelines for our audio-driven avatar generation framework.
\end{itemize}

\section{LLM Usage Statement}
We use LLMs (e.g., Gemini-2.5 and GPT-5) to polish our paragraphs.

\section{Additional Experimental Results}
\label{sec:additional_exp}

\subsection{Extending Autoregressive Generation to 10,000 Seconds}
To rigorously evaluate the long-horizon autoregressive capability of our model, we construct a stress test far exceeding the temporal range seen during training. Although the model is trained exclusively on 5-second clips---and its RoPE positions during training are randomly shifted only within a short-range window of a few minutes---we extend inference to an extreme 10,000-second horizon. Each audio input in GenBench-LongVideo ($\approx 7$ minutes per sample) is looped to fill the full 10,000-second duration, ensuring continuous and valid audio conditioning throughout the sequence while avoiding silence gaps. The model then performs fully autoregressive, block-wise generation following Self-Forcing~\cite{huang2025self}, relying entirely on accumulated KV caches and our long-horizon stability strategies (History Corrupt, AAS, and Rolling RoPE) throughout the 10,000-second rollout.

\begin{figure*}[t]
  \centering
   \includegraphics[width=0.88\linewidth]{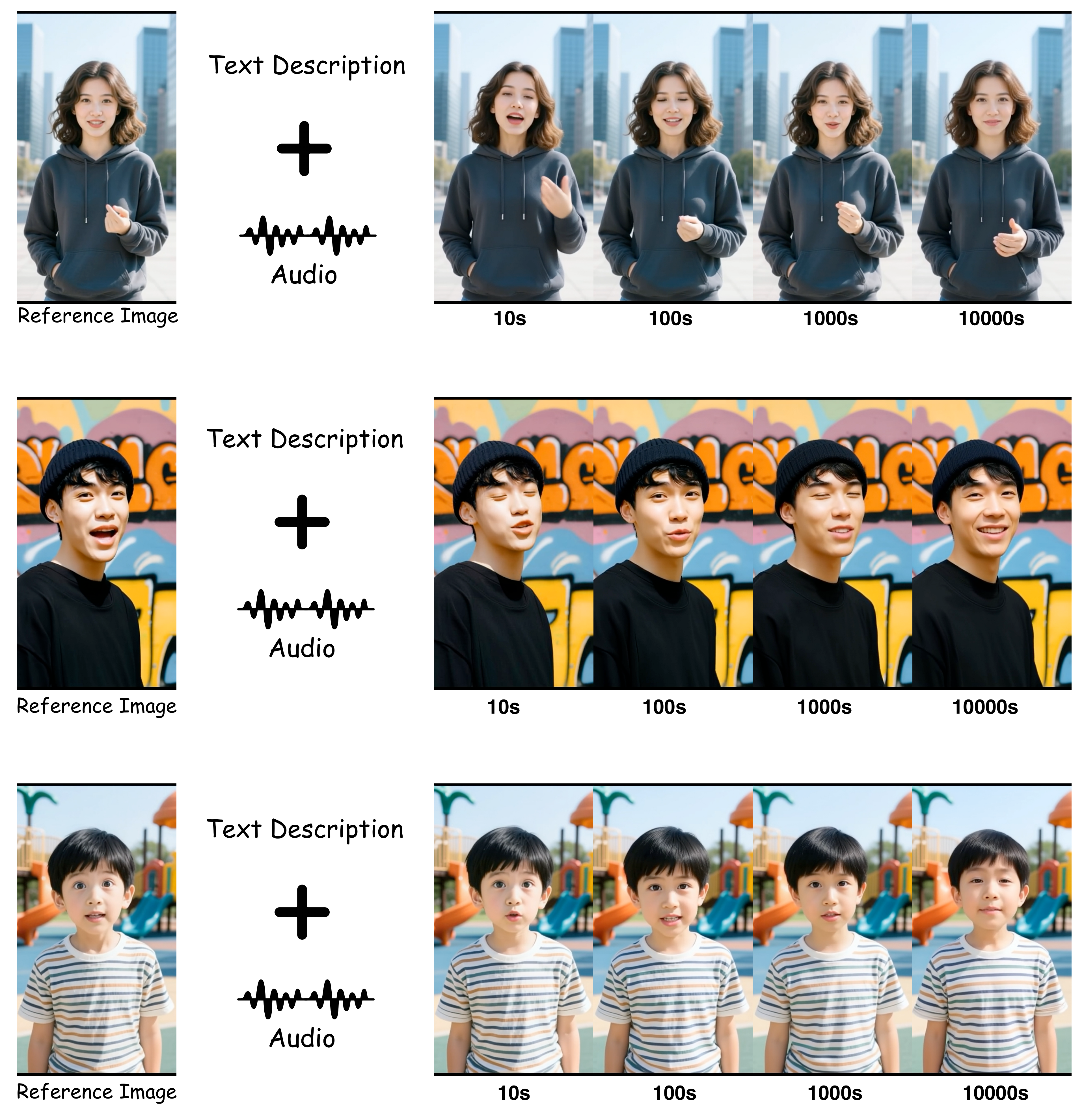}
   \caption{Visualization of the generated video at 10\,s, 100\,s, 1000\,s, and 10000\,s, demonstrating the model's strong capability in long-horizon temporal extrapolation.}
   \label{fig:visual1}
\end{figure*}

This setup intentionally exposes the model to RoPE indices over two orders of magnitude larger than those encountered in training (10k seconds corresponds to RoPE positions around 40k), a regime where existing methods typically suffer severe attention degradation, ID drift, or visual collapse. Self-Forcing++~\cite{cui2025self} demonstrates video generation of roughly 4 minutes, representing the longest horizon reported in prior work. In contrast, our model shows no observable quality decay or identity instability over the entire 10,000-second sequence. As shown in Table~\ref{tab:supp3}, perceptual quality (ASE, IQA), audio--visual synchronization (Sync-C), and semantic consistency (Dino-S) remain nearly unchanged across segments sampled at 0--10\,s, 100--110\,s, 1000--1010\,s, and 10000--10010\,s. Figure~\ref{fig:visual1} provides a representative case, demonstrating consistent identity and visual fidelity even at the 10k-second horizon.

Together, these results indicate that our long-video generation strategies---History Corrupt, Adaptive Attention Sink (AAS), and Rolling RoPE---allow the model to stably extrapolate far beyond its training regime, achieving an unprecedented 10,000-second autoregressive rollout without quality degradation.

\begin{table}[htbp]
\centering
\footnotesize
\caption{Evaluation of long-horizon temporal extrapolation at different time segments.}
\label{tab:supp3}
\begin{tabular}{l c c c c}
\toprule
 \multicolumn{1}{c}{Methods} & \multicolumn{4}{c}{Metrics}\\
 \cmidrule(lr){2-5}
 & ASE $\uparrow$ & IQA $\uparrow$ & Sync-C$\uparrow$ &Dino-S $\uparrow$\\
\midrule
 0-10s& 3.41& \textbf{4.77}& 7.10&\textbf{0.97}\\
  100-110s& \textbf{3.43}& 4.75& \textbf{7.22}&0.96\\
  1000-1010s& 3.40& 4.73& 6.98&0.96\\
  10000-10010s& 3.42& 4.76& 7.14&0.96\\
\bottomrule
\end{tabular}
\end{table}

% \begin{table*}[htbp]
% \centering
% \caption{Ablation Study on Long Video Generation.}
% \label{tab:ablation_long}
% \begin{tabular}{l c c c cl}
% \toprule
% \multicolumn{1}{c}{Methods} &\multicolumn{4}{c}{Metrics} & \\
% \cmidrule(lr){2-5}
%  & &ASE $\uparrow$ & IQA $\uparrow$ & Sync-C$\uparrow$ & Sync-D$\downarrow$   &\\
%  \midrule
%  & w/o Sink Frame&& & &  &\\
%  & w/o Sink Frame, w/ Reference Frame&& & &  &\\
%  & w/o RAFM&& & &  &\\
%  & w/o History Corrupt&& & &  &\\
%  & Ours && & &  &\\
% \bottomrule
% \end{tabular}
% \end{table*}

\subsection{Additional Comparison with Existing Methods}

Although we have already provided comprehensive comparisons on GenBench in the main paper, we further evaluate the robustness of our method within its training domain, AV-Speech. Specifically, we hold out 5\% of the original training videos and randomly sample 50 clips (5--10 seconds each) from this subset as an unseen test set. We report the same metrics used in the main evaluation; additionally, since ground-truth videos are available for this test set, we include FID and FVD to assess distribution alignment more thoroughly. The results are presented in Table~\ref{tab:supp2}.

The results in Table~\ref{tab:supp2} show that our method achieves competitive or superior performance across most metrics on the in-domain AV-Speech evaluation. Notably, OmniAvatar is also trained on AV-Speech and therefore serves as a strong in-domain baseline, yet our method remains competitive under this setting. Together with the results reported on multiple public benchmarks in the main paper, this additional experiment verifies that our approach performs reliably within its training domain and alleviates potential concerns about relying solely on benchmark-specific evaluations.
\begin{table*}[htbp]
\centering
\small
\caption{Quantitative comparisons on the in-domain AV-Speech test set.}
\label{tab:supp2}
\resizebox{\linewidth}{!}{%
\begin{tabular}{l c c c c c c c c c}
\toprule
\multicolumn{1}{c}{Dataset} & \multicolumn{1}{c}{Model} &  \multicolumn{8}{c}{Metrics}\\
 \cmidrule(lr){3-10}& &  FID$\downarrow$& FVD$\downarrow$&ASE $\uparrow$ & IQA $\uparrow$ & Sync-C$\uparrow$ & Sync-D$\downarrow$  &Dino-S $\uparrow$& FPS $\uparrow$ \\
\midrule
\multirow{7}{*}{AV-Speech}
 & Ditto\cite{li2024ditto} &\textbf{46.27}&660.01 &2.21 & 3.75 & 4.84 & 9.05  &\textbf{0.99} & 21.80 \\
 & Echomimic-V2\cite{meng2025echomimicv2strikingsimplifiedsemibody} &176.74  &2059.81 &1.88 & 3.29 & 4.07 & 9.38  &0.78 & 0.53 \\
 & Hallo3\cite{cui2025hallo3} &138.40  &1412.93 &1.87 & 3.35 & 4.50 & 9.99  &0.93 & 0.26 \\
 & StableAvatar\cite{tu2025stableavatar} &98.32  &730.12 &2.14 & 3.55 & 5.72 & 9.01  &0.93 & 0.64 \\
 & OmniAvatar\cite{gan2025omniavatar} &50.42&570.32 &2.16 & 3.68 & 6.04& 8.37&0.95& 0.16 \\
 & WanS2V\cite{gao2025wans2vaudiodrivencinematicvideo} &73.68  &642.48 &2.20 & 3.71 & 4.90 & 9.02  &0.95& 0.25 \\
 & Ours &  48.91& \textbf{502.37}&\textbf{2.30}& \textbf{3.88}& \textbf{6.21}& \textbf{8.31}&0.95& \textbf{45.2}\\
 \bottomrule
\end{tabular}%
}
\end{table*}

\subsection{User Study}
\begin{figure}[t]
  \centering
  \includegraphics[width=\linewidth]{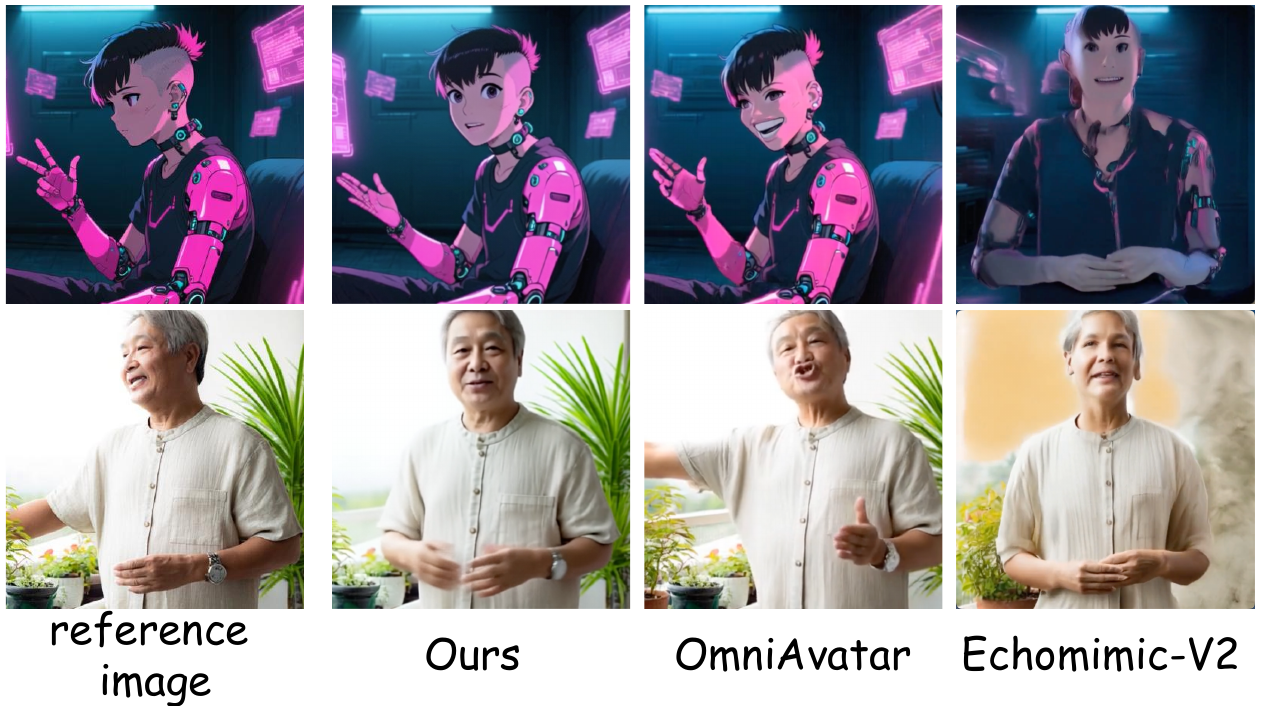}
  \caption{Qualitative examples from the user study. OmniAvatar exhibits exaggerated motion with degraded identity fidelity, while EchoMimic V2 ignores the reference pose and produces a fixed body template, leading to visible distortion.}
  \label{fig:user-study}
\end{figure}

Prior work such as THEval~\cite{quignon2025thevalevaluationframeworktalking} has shown that popular metrics for talking-avatar evaluation (e.g., Sync-C) often diverge from human perception, as models can exploit them by exaggerating lip motion.
To bridge this gap, we conduct a double-blind user study with 20 participants, who rate generated videos from all methods on three perceptual dimensions: Naturalness, Synchronization, and Consistency. Here, Synchronization refers to the holistic audiovisual coherence~\cite{wang2025fantasytalking}---including facial expressions, gestures, and posture transitions---rather than the frame-level lip alignment measured by Sync-C.
The final scores are averaged across participants and normalized to a 0--100 scale for comparison.
As summarized in Table~\ref{tab:user_study}, our method attains the highest Synchronization and Consistency scores. WanS2V achieves the best Naturalness, likely because its non-distilled diffusion backbone preserves smoother motion dynamics; we observe that our distilled model introduces slightly accelerated motion tempo, which marginally affects perceived naturalness but does not compromise synchronization or identity consistency.
Figure~\ref{fig:user-study} provides representative qualitative examples. OmniAvatar, despite achieving strong objective metrics, produces exaggerated motions that compromise identity preservation over time, leading to lower Naturalness ratings from human evaluators. EchoMimic V2, which relies on fixed hand-landmark templates, tends to ignore the body pose present in the reference image and instead generates a fixed default posture, causing severe visual distortion when the input depicts non-standard poses.
\begin{table}[htbp]
\centering
\footnotesize
\caption{User study results on perceptual evaluation (higher is better). Each score denotes the mean normalized user rating.}
\label{tab:user_study}
\begin{tabular}{lccc}
\toprule
\multicolumn{1}{c}{Model} & Naturalness $\uparrow$ & Synchronization $\uparrow$ & Consistency $\uparrow$ \\
\midrule
Ditto~\cite{li2024ditto} & 78.2 & 40.5& 90.2\\
EchoMimic-V2~\cite{meng2025echomimicv2strikingsimplifiedsemibody} & 60.3 & 71.1& 38.7\\
Hallo3~\cite{cui2025hallo3} & 78.5& 69.2 & 89.3\\
StableAvatar~\cite{tu2025stableavatar} & 68.7 & 70.8& 88.9\\
OmniAvatar~\cite{gan2025omniavatar} & 71.1& 78.5& 90.8\\
WanS2V~\cite{gao2025wans2vaudiodrivencinematicvideo} & \textbf{84.3}& 85.2& 92.0\\
Ours & 80.1& \textbf{86.0}& \textbf{93.2}\\
\bottomrule
\end{tabular}
\end{table}

% \begin{table}[htbp]
% \centering
% \footnotesize
% \caption{User study results on perceptual evaluation (higher is better). Each score denotes the mean normalized user rating (5-point scale).}
% \label{tab:user_study}
% \begin{tabular}{lccc}
% \toprule
% \multicolumn{1}{c}{Model} & Naturalness $\uparrow$ & Synchronization $\uparrow$ & Consistency $\uparrow$ \\
% \midrule
% Ditto~\cite{li2024ditto} & 3.91 & 2.03 & 4.51 \\
% EchoMimic-V2~\cite{meng2025echomimicv2strikingsimplifiedsemibody} & 3.02 & 3.56 & 1.94 \\
% Hallo3~\cite{cui2025hallo3} & 3.93 & 3.46 & 4.47 \\
% StableAvatar~\cite{tu2025stableavatar} & 3.44 & 3.54 & 4.45 \\
% OmniAvatar~\cite{gan2025omniavatar} & 3.56 & 3.93 & 4.54 \\
% WanS2V~\cite{gao2025wans2vaudiodrivencinematicvideo} & 4.22 & \textbf{4.26} & \textbf{4.60} \\
% Ours & \textbf{4.32} & 4.03 & 4.56 \\
% \bottomrule
% \end{tabular}
% \end{table}

\subsection{Effect of Block Size}
\label{sec:block_size}

\begin{table}[htbp]
\centering
\footnotesize
\caption{Effect of block size (number of latent frames per block) on GenBench-ShortVideo. All variants are trained end-to-end (Stage~1 + Stage~2) with identical hyperparameters.}
\label{tab:block_size}
\begin{tabular}{l c c c c c c}
\toprule
\multicolumn{1}{c}{Block Size} & ASE $\uparrow$ & IQA $\uparrow$ & Sync-C $\uparrow$ & Sync-D $\downarrow$ & Dino-S $\uparrow$ & TTFF (s) $\downarrow$ \\
\midrule
1 latent  & 3.30  & 4.35  & 4.12  & 10.41 & 0.96 & \textbf{0.40} \\
2 latents & 3.36  & 4.41  & 5.38  & 9.52  & 0.96 & 0.67 \\
3 latents & \textbf{3.44}  & 4.49  & \textbf{7.03}  & 8.30  & \textbf{0.96} & 0.94 \\
4 latents & 3.43  & \textbf{4.51}  & 6.98  & \textbf{8.26}  & \textbf{0.96} & 1.21 \\
\bottomrule
\end{tabular}
\end{table}

We ablate the number of latent frames per block while keeping all other training and inference hyperparameters fixed. As shown in Table~\ref{tab:block_size}, reducing the block size to 1 or 2 latents leads to a pronounced drop in audio--visual synchronization (Sync-C drops from 7.03 to 4.12 and 5.38, respectively), while visual quality metrics (ASE, IQA) degrade more mildly and identity consistency (Dino-S) remains essentially unchanged. Qualitatively, we observe that smaller block sizes produce videos with severely diminished motion dynamics---the generated avatar exhibits near-static facial expressions and minimal gesture variation---consistent with findings in CausVid~\cite{yin2025slow}, where single-frame autoregressive generation suffers from limited temporal expressiveness. Increasing the block size to 4 yields performance on par with 3---block size 4 slightly edges ahead on IQA and Sync-D, while block size 3 is marginally better on ASE and Sync-C---but incurs a noticeably higher TTFF (1.21\,s vs.\ 0.94\,s). We therefore adopt a block size of 3 as the default, which provides the best trade-off between generation quality and interactive latency.

\section{Additional Evaluation Details}
\label{sec:additional setting}
\textbf{Inference Configuration.} All methods are evaluated on a single H100 node under identical hardware conditions.
To ensure fair comparison, we utilize multi-GPU parallel inference for all methods where the official open-source code provides appropriate scripts. For methods that lack support for parallel acceleration, specifically EchoMimic V2 and Ditto, inference is conducted on a single H100 GPU. Regarding resolution, we use a fixed resolution of $720\times400$ for all models except Hallo3, which does not support arbitrary aspect ratios or resolutions. For Hallo3, we crop both the input and the ground-truth frames to $512\times512$. Furthermore, EchoMimic V2 is excluded from the long-video comparative experiments, as its reliance on per-frame skeleton templates prevents it from effectively performing long-duration inference.

\textbf{Runtime Metrics.} For runtime evaluation, we report two key efficiency metrics. \textbf{FPS (Frames Per Second)} is measured from the moment the inference pipeline is initialized and includes the full end-to-end cost: the diffusion model's denoising time, VAE decoding time, and any additional CPU-side processing. \textbf{Time-to-First-Frame (TTFF)} accounts for the total latency from audio signal arrival to visual output, calculated as the sum of (i) the random arrival latency---the waiting time between the arrival of an audio interaction signal and the next frame boundary---followed by (ii) the full denoising latency of the first frame and (iii) its VAE decoding cost. Note that random arrival latency depends on the output frame rate, and thus TTFF is inherently coupled with the FPS of the method.
All FPS measurements for all methods are evaluated on the GenBench-LongVideo benchmark, where long-sequence testing provides more stable and accurate runtime estimation.

\section{Kernel-Level Optimizations}
\label{sec:kernel_opt}
This section details the kernel-level optimizations referred to as \textit{Kernel Opt.}\ in the efficiency ablation study of the main paper (Sec.~4.2). To enable real-time streaming inference with a 14B-parameter diffusion transformer, we apply a set of optimizations spanning quantization, attention kernels, graph compilation, weight merging, and VAE decoding.

\textbf{FP8 Quantization.}
All linear projections in the DiT are quantized to FP8 (E4M3) with per-tensor dynamic scaling. Embedding layers, the output head, and the audio encoder's final projection are kept in FP16 to preserve numerical stability. This reduces per-GPU VRAM from ${\sim}$80\,GB to under 48\,GB, enabling deployment on single 48\,GB GPUs.
To verify that FP8 quantization does not degrade generation quality, we compare the quantized model against its FP16 counterpart on both GenBench subsets (Table~\ref{tab:fp8_ablation}). Disabling FP8 yields only marginal quality differences---IQA improves slightly while other metrics remain virtually unchanged---yet throughput drops from 45.2 to 36.0\,FPS. We therefore adopt FP8 as the default, and all metrics reported in the main paper use this configuration.

\begin{table}[h]
\centering
\footnotesize
\caption{Effect of FP8 quantization on generation quality and throughput.}
\label{tab:fp8_ablation}
\setlength{\tabcolsep}{3pt}
\resizebox{\columnwidth}{!}{%
\begin{tabular}{l l c c c c c c}
\toprule
\multicolumn{1}{c}{Dataset} & \multicolumn{1}{c}{Setting} & ASE$\uparrow$ & IQA$\uparrow$ & Sync-C$\uparrow$ & Sync-D$\downarrow$ & Dino-S$\uparrow$ & FPS$\uparrow$ \\
\midrule
\multirow{2}{*}{GenBench-Short}
 & w/o FP8 (FP16) & \textbf{3.45} & \textbf{4.55} & 7.02 & \textbf{8.28} & \textbf{0.96} & 36.0 \\
 & Ours (FP8)      & 3.44 & 4.51 & \textbf{7.03} & 8.30 & \textbf{0.96} & \textbf{45.2} \\
\midrule
\multirow{2}{*}{GenBench-Long}
 & w/o FP8 (FP16) & \textbf{3.42} & \textbf{4.79} & \textbf{7.16} & \textbf{8.30} & \textbf{0.97} & 36.0 \\
 & Ours (FP8)      & \textbf{3.42} & 4.76 & \textbf{7.16} & 8.31 & \textbf{0.97} & \textbf{45.2} \\
\bottomrule
\end{tabular}%
}
\end{table}

\textbf{FlashAttention-3 and cuDNN Fused Attention.}
We adopt a tiered attention dispatch strategy. On Hopper-class GPUs, FlashAttention-3 with variable-length sequence support serves as the primary kernel, efficiently handling the concatenation of cached KV entries and current tokens in streaming inference. FlashAttention-2 is used as a fallback on older architectures. Where applicable, we further dispatch to a fused cuDNN scaled dot-product kernel that computes the entire QKV attention in a single pass, avoiding materialization of the full attention matrix.

\textbf{Graph Compilation.}
We apply PyTorch's ahead-of-time graph compilation to the DiT forward pass, RoPE computation, and streaming VAE decoding. This enables automatic operator fusion (e.g., fusing RoPE element-wise operations, layer-norm with subsequent linear projections) and kernel auto-tuning across the denoising loop.

\textbf{LoRA Weight Merging.}
LoRA adapters trained during distillation are permanently merged into the base model weights before inference, eliminating the runtime overhead of maintaining and computing through separate low-rank branches.

\textbf{Streaming VAE Feature Caching.}
We implement a causal VAE decoder that processes latents frame-by-frame while maintaining a temporal feature cache: each causal convolution layer retains the activations of the most recent 2 frames, which are prepended as causal context when decoding the next frame, enabling incremental decoding without re-computation.
In the multi-GPU TPP configuration, a dedicated GPU runs the streaming VAE in parallel with the DiT---decoded blocks are sent via point-to-point communication as soon as they are produced, effectively hiding VAE latency behind DiT compute.

\textbf{Model Offloading.}
For single-GPU deployment, the text encoder, audio encoder, and VAE are offloaded to CPU after their respective encoding phases, freeing VRAM for the DiT denoising loop. KV caches can optionally be offloaded to CPU between clips to further reduce peak memory.

Together, these optimizations yield ${\sim}$2.5$\times$ peak and ${\sim}$3$\times$ average FPS improvement over the unoptimized baseline, achieving stable 45+ FPS real-time streaming on 5$\times$H100 with 4-step sampling. On a more cost-effective 5$\times$H20 configuration, the system still delivers 18 FPS end-to-end with a TTFF of 3.12\,s, demonstrating practical deployability beyond high-end datacenter hardware.

\section{Algorithm Details}
\label{sec:algo}

\begin{figure*}[t!]
  \centering
   \includegraphics[width=0.7\linewidth]{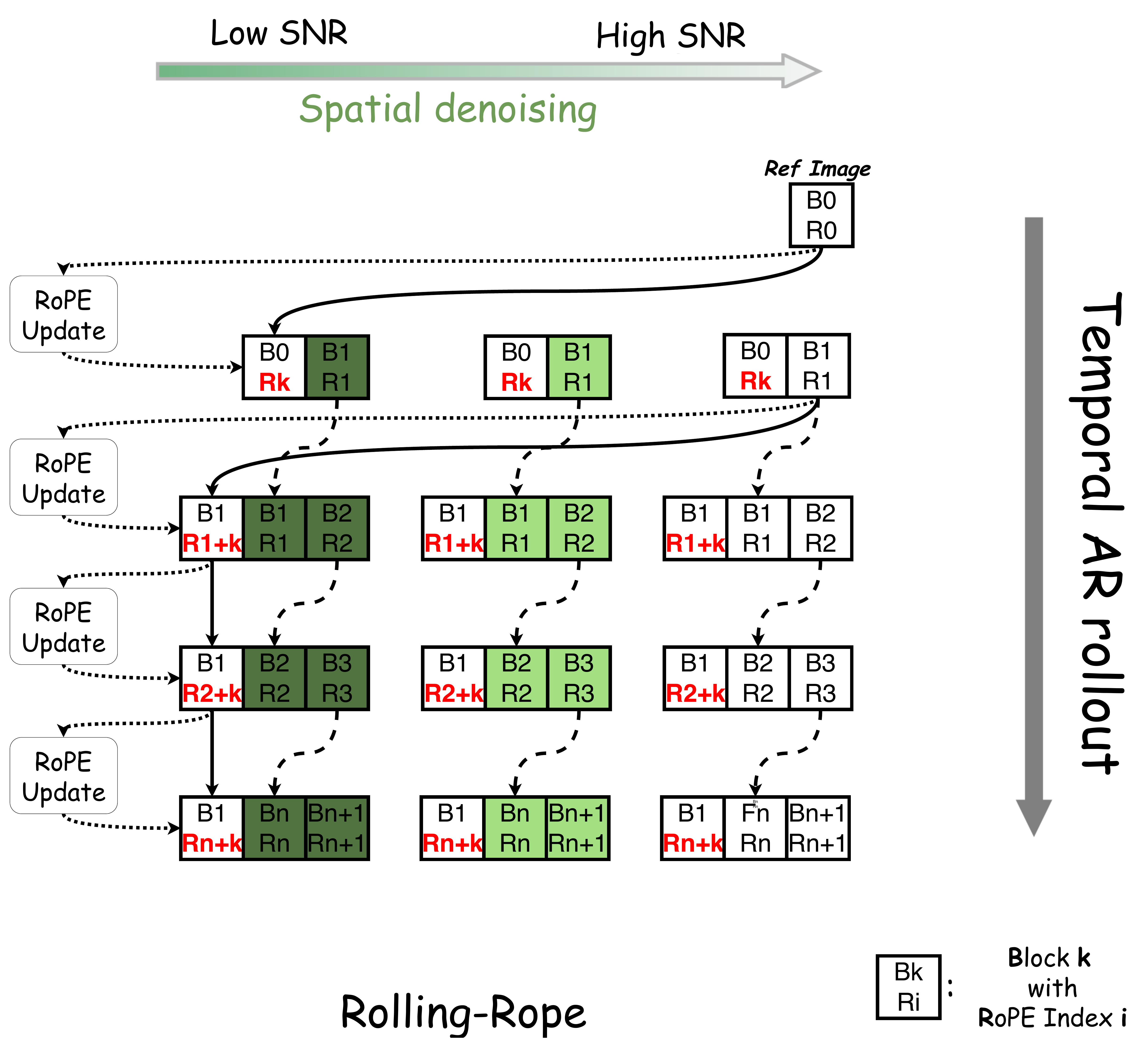}
   \caption{Visualization of the proposed Rolling-RoPE mechanism. Horizontally, each row follows the spatial denoising order from low to high SNR; vertically, each column shows the autoregressive rollout over frames. Each small \textbf{rectangle} denotes the latent of a block, and the \textbf{number} inside represents its block index and its RoPE index, respectively. \textbf{Red} marks indicate the components modified relative to the baseline.
   \textbf{Solid arrows} denote sink-frame passing, \textbf{sparse dashed arrows} denote standard KV-cache passing, and \textbf{dense dashed arrows} indicate RoPE updates, where each block is reassigned updated positional embeddings. Rolling-RoPE dynamically \textbf{increases the RoPE index} of the sink frame throughout AR rollout, keeping the sink frame's RoPE index slightly larger than that of the current noisy block, ensuring a stable and appropriate relative positional distance throughout AR rollout.}
   \label{fig:method4}
\end{figure*}

\begin{figure*}[t!]
  \centering
   \includegraphics[width=0.88\linewidth]{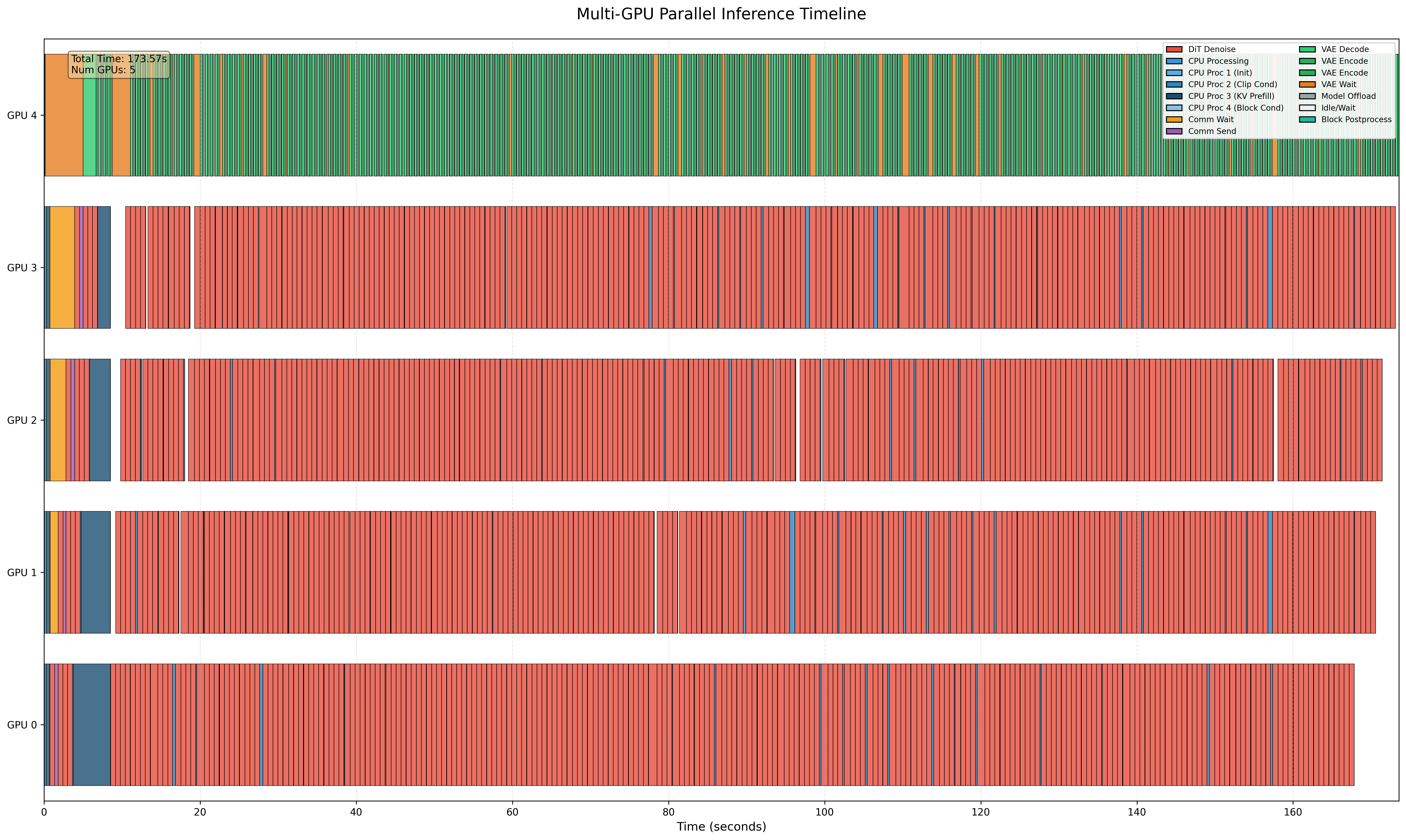}
   \caption{Multi-GPU Parallel Inference Timeline. This chart visualizes the computation and waiting periods for each GPU. 
   The two distinct \textbf{white} gaps on the left represent the initial warmup phase (including the secondary warmup for AAS).
   Following these, the majority of the processing time is dedicated to DiT computation (shown in \textbf{red}), reflecting high utilization and stable frame rates. 
   Sporadic \textbf{white} gaps (idle time) appearing thereafter are present due to rate fluctuations, but their rarity ensures a negligible impact on performance.}
   \label{fig:TPP}
\end{figure*}

We provide detailed pseudocode for completeness and reproducibility.

\textbf{Training.} Our two-stage training framework is described as follows. In Stage~1 (Diffusion Forcing Pretraining), we apply block-wise independent noise scheduling and causal attention, with motion frames serving as scaffolding to bootstrap long-horizon temporal modeling (see main text Sec.~3.2). In Stage~2 (Self-Forcing Distillation), the motion frames are removed and replaced by the rolling KV cache. As shown in Algorithm~\ref{alg:self_forcing1}, the self-forcing DMD training follows \cite{huang2025self} but removes the additional forward pass used to refresh the KV cache after denoising. This ensures that the model is consistently trained with noisy KV states, aligning the training process with the actual autoregressive inference and improving robustness to error accumulation---a strategy we call \textbf{History Corrupt}. Due to the substantial memory footprint of DMD training, we additionally implement a lightweight memory-reduction strategy using \textbf{block-wise gradient accumulation} (Algorithm~\ref{alg:self_forcing2}), which partitions the backward graph by blocks and accumulates the resulting gradients across multiple steps, preserving training behavior while significantly reducing peak memory usage and enabling even single-node H100 training.

\textbf{Inference.} The single-GPU inference procedure is provided in Algorithm~\ref{alg:inference1}. It builds upon the rolling-KV-cache inference algorithm from \cite{huang2025self} with the addition of AAS and Rolling RoPE. Below we detail the concrete implementation of these two mechanisms.

\textbf{Adaptive Attention Sink (AAS).} As described in the main text, the user-provided reference image initially serves as the sink frame. To eliminate the distributional mismatch between real conditioning and model-generated content, AAS replaces the sink frame with the model's own first generated \emph{latent} (i.e., the denoised output of the first block) immediately after the first block is produced. Crucially, this replacement operates entirely in latent space---no additional VAE encoding or decoding is required---keeping the sink frame on the model's generation manifold with negligible overhead. The updated sink frame then persists as the sole identity anchor for all subsequent blocks (see Algorithm~\ref{alg:inference1}, line 16--17; Algorithm~\ref{alg:inference2}, line 13--14 and 18--19).

\textbf{Rolling RoPE.} During standard autoregressive rollout, the sink frame's RoPE index remains fixed at position 0 while the generated blocks' indices grow monotonically (e.g., 1, 2, 3, \ldots). As generation progresses, the relative positional distance between the sink frame and the current block becomes arbitrarily large---far exceeding positions seen during training---causing attention to the sink frame to degrade. Rolling RoPE addresses this by dynamically reassigning the sink frame's RoPE index at every block step: specifically, the sink frame's index is set to be slightly ahead of the current noisy block's index (within the training-time offset range), ensuring that the relative positional distance remains bounded and consistent with training throughout the entire rollout. This update is applied to the sink frame's cached KV entries by recomputing RoPE embeddings in place, denoted as $\Phi(\SinkFrame)$ in Algorithms~\ref{alg:inference1} and~\ref{alg:inference2}. Figure~\ref{fig:method4} illustrates the RoPE update performed during autoregressive generation, where the sink frame is continuously re-assigned with updated positional embeddings.

The multi-GPU TPP inference is detailed in Algorithm~\ref{alg:inference2}, which outlines the parallel execution procedure with minimal computation overlap and communication overhead. Figure~\ref{fig:TPP} further visualizes the computation and waiting time of each GPU. After the initial warmup (including a second warmup required for AAS), the majority of each GPU's time is devoted to DiT computation (red), demonstrating high utilization and stable frame rates.

      \begin{algorithm}[H]
        \caption{Self-Forcing DMD with History Corrupt}
        \small
        \begin{algorithmic}[1]
          \Require Timesteps $\{t_1, \dots, t_T\}$
          \Require Number of video frames $N$
          \Require Conditions of $N$ frames $C_{1:N}$ (including audio,text,ref image)
          \Require Generator $G_\theta$ (returns KV embeddings via $G_\theta^{\mathrm{KV}}$)
          \Loop
            \State Initialize model output $\ModelOuput \gets []$
            \State Initialize KV cache $\KVSet \gets []$
            \State Sample $s \sim \text{Uniform}(1, 2, \ldots, T)$
            \For{$i = 1, \dots, N$}
              \State Initialize $x^i_{t_T} \sim \mathcal{N}(0, I)$
              \For{$j = T, \dots, s$}
                \If{$j = s$}
                  \State Enable gradient computation
                  \State Set $\KVOutput^i$, $\hat{x}^i_{0} \gets G_\theta^\text{KV}(x^i_{t_{j}}; t_j, \KVSet,C_i)$
                  \State $\ModelOuput{\texttt{.append}}(\hat x^i_{0})$
                  \State Detach $\KVOutput^i$ from gradient graph
                  \State $\KVSet{\texttt{.append}}(\KVOutput^i)$  \Comment{Noisy KV Cache}
                \Else
                  \State Disable gradient computation
                  \State Set $\hat{x}^i_{0} \gets G_\theta(x^i_{t_j}; t_j, \KVSet,C_i)$
                  \State Sample $\epsilon \sim \mathcal{N}(0, I)$
                  \State Set $x^i_{t_{j-1}} \gets \Psi(\hat{x}^i_0, \epsilon, t_{j-1})$
                \EndIf
              \EndFor
            \EndFor
            \State Update $\theta$ via distribution matching loss
          \EndLoop
        \end{algorithmic}
        \label{alg:self_forcing1}
      \end{algorithm}
    %   \begin{algorithm}[H]
    %     \caption{Self-Forcing Distribution Matching Distillation with History Corrupt}
    %     \small
    %     \begin{algorithmic}[1]
    %       \Require Denoise timesteps $\{t_1, \dots, t_T\}$
    %       \Require Number of video frames $N$
    %       \Require AR diffusion model $G_\theta$ (returns KV embeddings via $G_\theta^{\mathrm{KV}}$)
    %       \Loop
    %         \State Initialize model output $\ModelOuput \gets []$
    %         \State Initialize KV cache $\KVSet \gets []$
    %         \State Sample $s \sim \text{Uniform}(1, 2, \ldots, T)$
    %         \For{$i = 1, \dots, N$}
    %           \State Initialize $x^i_{t_T} \sim \mathcal{N}(0, I)$
    %           \For{$j = T, \dots, s$}
    %             \If{$j = s$}
    %               \State Enable gradient computation
    %               \State Set $\KVOutput^i$, $\hat{x}^i_{0} \gets G_\theta^\text{KV}(x^i_{t_{j}}; t_j, \KVSet)$
    %               \State $\ModelOuput{\texttt{.append}}(\hat x^i_{0})$
    %               \State Detach $\KVOutput^i$ from gradient graph
    %               \State $\KVSet{\texttt{.append}}(\KVOutput^i)$
    %             \Else
    %               \State Disable gradient computation
    %               \State Set $\hat{x}^i_{0} \gets G_\theta(x^i_{t_j}; t_j, \KVSet)$
    %               \State Sample $\epsilon \sim \mathcal{N}(0, I)$
    %               \State Set $x^i_{t_{j-1}} \gets \Psi(\hat{x}^i_0, \epsilon, t_{j-1})$
    %             \EndIf
    %           \EndFor
    %         \EndFor
    %         \State Update $\theta$ via distribution matching loss
    %       \EndLoop
    %     \end{algorithmic}
    %     \label{alg:self_forcing}
    %   \end{algorithm}
    % \end{minipage}
    \hfill
    % \begin{minipage}[t]{0.48\textwidth}
      \begin{algorithm}[H]
        \caption{Self-Forcing DMD with History Corrupt and Block-wise Gradient Accumulation}
        \small
        \setcounter{ALG@line}{0}
        \begin{algorithmic}[1]
          \Require Timesteps $\{t_1, \dots, t_T\}$
          \Require Number of video frames $N$
          \Require Conditions of $N$ frames $C_{1:N}$ (including audio,text,ref image)
          \Require Generator $G_\theta$ (extra returns KV embeddings via $G_\theta^{\mathrm{KV}}$)
          \Loop
            \State Initialize model output $\ModelOuput \gets []$
            \State Initialize noisy latent cache $\NoisyLatent \gets []$
            \State Initialize KV cache $\KVSet \gets []$
            \State Sample $s \sim \text{Uniform}(1, 2, \ldots, T)$
            \State Disable gradient computation
            \For{$i = 1, \dots, N$}
              \State Initialize $x^i_{t_T} \sim \mathcal{N}(0, I)$
              \For{$j = T, \dots, s$}
                \If{$j = s$}
    
                  \State Set $\KVOutput^i$, $\hat{x}^i_{0} \gets G_\theta^\text{KV}(x^i_{t_{j}}; t_j, \KVSet,C_i)$
                  \State Detach $x^i_{t_{j}},\hat{x}^i_{0},\KVOutput^i$ from gradient graph
                % \State Detach $\hat{x}^i_{0}$ from gradient graph
                %   \State Detach $\KVOutput^i$ from gradient graph
                  \State $\NoisyLatent{\texttt{.append}}(x^i_{t_{j}})$
                \State $\ModelOuput{\texttt{.append}}(\hat x^i_{0})$
                  \State $\KVSet{\texttt{.append}}(\KVOutput^i)$
                \Else
                  \State Set $\hat{x}^i_{0} \gets G_\theta(x^i_{t_j}; t_j, \KVSet,C_i)$
                  \State Sample $\epsilon \sim \mathcal{N}(0, I)$
                  \State Set $x^i_{t_{j-1}} \gets \Psi(\hat{x}^i_0, \epsilon, t_{j-1})$
                \EndIf
              \EndFor
            \EndFor
            \For{$i = N, \dots, 1$}
            \State Initialize temporary model output $\ModelOuputTemp \gets []$
            \For{$j = 1, \dots, N$}
             \If{$i=j$}
                \State Set $x^i_{t_{s}} \gets \NoisyLatent[j]$
                 \State Enable gradient computation
                  \State Set $\hat{x}^i_{0} \gets G_\theta(x^i_{t_{s}}; t_s, \KVSet,C_i)$
                  \State $\ModelOuputTemp{\texttt{.append}}(\hat x^i_{0})$ \Comment{Handle Partial Gradient}
                  \State Disable gradient computation
             \Else
            \State $\ModelOuputTemp{\texttt{.append}}(\ModelOuput[j])$
              \EndIf
            \EndFor
              \State Accumulate gradient of $\theta$ via DMD loss
              \State $\KVSet$.pop(i) \Comment{Free Memory}
            \EndFor
            \State Update $\theta$ 
          \EndLoop
        \end{algorithmic}
        \label{alg:self_forcing2}
      \end{algorithm}
    % \end{minipage}
    % \clearpage
    % \hfill
    % % \begin{minipage}[t]{0.48\textwidth}
    %   \begin{algorithm}[H]
    %     \caption{Autoregressive Diffusion Inference with Rolling Rope}
    %     \small
    %     \begin{algorithmic}[1]
    %       \Require KV cache of size $L$ frames
    %       \Require Denoise timesteps $\{t_1, \dots, t_T\}$
    %       \Require Number of generated frames $M$
    %       \Require AR diffusion model $G_\theta$ (returns KV embeddings via $G_\theta^\text{KV}$)
    %       \State Initialize model output $\ModelOuput \gets []$
    %       \State Initialize KV cache $\KVSet \gets []$
    %       \For{$i = 1, \dots, M$}
    %         \State Initialize $x^i_{t_T} \sim \mathcal{N}(0, I)$
    %         \For{$j = T, \dots, 1$}
    %           \State Set $\hat{x}^i_{0} \gets G_\theta(x^i_{t_j}; t_j, \KVSet)$
    %           \If{$j = 1$}
    %             \State $\ModelOuput{\texttt{.append}}(\hat x^i_{0})$
    %             \State Cache $\KVOutput^i \gets G_\theta^\text{KV}(\hat{x}^i_{0}; 0, \KVSet)$
    %             \If{$|\KVSet| = L$}
    %               \State $\KVSet.\mathrm{pop}(0)$ \Comment{Cache eviction}
    %             \EndIf
    %             \State $\KVSet{\texttt{.append}}(\KVOutput^i)$
    %           \Else
    %             \State Sample $\epsilon \sim \mathcal{N}(0, I)$
    %             \State Set $x^i_{t_{j-1}} \gets \Psi(\hat{x}^i_0, \epsilon, t_{j-1})$
    %           \EndIf
    %         \EndFor
    %       \EndFor
    %       \State \Return $\ModelOuput$
    %     \end{algorithmic}
    %     \label{alg:inference}
    %   \end{algorithm}
    % % \end{minipage}
    % \vspace{-1em}
      \begin{algorithm}[H]
        \caption{Single-GPU AR Inference with AAS and Rolling RoPE}
        \small
        \setcounter{ALG@line}{0}
        \begin{algorithmic}[1]
          \Require Per-timestep KV caches, each with size $L$
          \Require Timesteps $\{t_1, \dots, t_T\}$
          \Require Number of generated frames $M$
          \Require Conditions of $N$ frames $C_{1:N}$ (including audio,text)
        \Require Ref image $R$ 
          \Require Flow-Matching Model $v_\theta^\text{KV}$ (extra returns KV embeddings)
          \Require Rolling RoPE transform $\Phi(\cdot)$ 
          \Require VAE Decoder $\text{VAE}(\cdot)$ 
          \State Initialize model output $\ModelOuput \gets []$
          \State Initialize KV caches $\{\KVSet_1, \dots, \KVSet_T\} \gets []$
          \State Initialize Rolling Sink Frame $ \SinkFrame \gets R$
          \State Initialize $dt \gets -1/T$
          \For{$i = 1, \dots, M$}
            \State Initialize $x^i \sim \mathcal{N}(0, I)$
            \For{$j = T, \dots, 1$}
              \State Set $\hat{v}^i_{j},\KVOutput^i_j \gets v_\theta^\text{KV}(x^i; t_j, \KVSet_j,C_i,\Phi(\SinkFrame))$ \Comment{RoPE Update}
              \State Set $x^i \gets x^i + \hat{v}^i_{j} dt$
            \If{$|\KVSet_j| = L$}
              \State $\KVSet_j.\mathrm{pop}(0)$ 
            \EndIf
            \State $\KVSet_j{\texttt{.append}}(\KVOutput^i_j)$
            \EndFor
            \State $\ModelOuput{\texttt{.append}}(  \text{VAE}(x^i))$
            \If{$i = 1$}
                \State $\SinkFrame \gets  x^i$  \Comment{AAS Update}
            \EndIf
          \EndFor
          \State \Return $\ModelOuput$
        \end{algorithmic}
        \label{alg:inference1}
      \end{algorithm}
    \begin{algorithm}[H]
      \caption{TPP with AAS and Rolling RoPE}
      \small
      \setcounter{ALG@line}{0}
      \begin{algorithmic}[1]
          \Require GPU Index $k$
        \Require Per-timestep KV caches, each with size $L$
        \Require Timesteps $\{t_1, \dots, t_T\}$
        \Require Number of generated frames $M$
        \Require Conditions of $N$ frames $C_{1:N}$ (including audio,text)
          \Require Ref image $R$ 
        \Require Flow-Matching Model $v_\theta^\text{KV}$ (extra returns KV embeddings)
        \Require Rolling RoPE transform $\Phi(\cdot)$ 
      \Require VAE Decoder $\text{VAE}(\cdot)$ 
        \State Initialize model output $\ModelOuput \gets []$
        \State Initialize KV cache $\KVSet \gets []$
        \State Initialize Rolling Sink Frame $ \SinkFrame \gets R$
        \State Initialize $dt \gets -1/T$
        \For{$i = 1, \dots, M$}
          \If{k=1}
          \State Initialize $x^i \sim \mathcal{N}(0, I)$
          \Else
          \State $x^i \gets $ dist.recv()
          \EndIf
          \If{k=T+1}   \Comment{VAE Device}
              \State $\ModelOuput{\texttt{.append}}( \text{VAE}(x^i))$
              \If{$i = 1$}
                  \State dist.broadcast$(x^i)$ \Comment{Broadcast AAS Latent}
              \EndIf
              \State \textbf{continue}
          \Else \Comment{DiT Device}
              \If{i=2}
              \State $\SinkFrame \gets $dist.recv()  \Comment{Update AAS}
              \EndIf
            \State Set $\hat{v}^i_{k},\KVOutput^i \gets v_\theta^\text{KV}(x^i; t_{T-k+1}, \KVSet,C_i,\Phi(\SinkFrame))$ 
          \If{$|\KVSet| = L$}
            \State $\KVSet.\mathrm{pop}(0)$ 
          \EndIf 
          \State $\KVSet{\texttt{.append}}(\KVOutput^i)$
          \State Set $x^i \gets x^i + \hat{v}^i_{k} dt$
          \State  dist.send($x^i$)
      \EndIf
      \EndFor
  
        \State \Return $\ModelOuput$
      \end{algorithmic}
      \label{alg:inference2}
    \end{algorithm}

\section{Additional Visual Results}
\label{sec:visuals}
We provide additional qualitative examples to further illustrate the model's long-horizon generation capability. As shown in Figure~\ref{fig:visual1}, our method maintains stable identity, consistent lip movements, and coherent visual appearance when extrapolating videos far beyond the training horizon. For three different subjects, the generated frames at 10\,s, 100\,s, 1000\,s, and 10000\,s remain visually aligned with the reference image and follow the audio-driven motion patterns without exhibiting temporal drift or degradation. These results highlight the robustness of our approach in producing coherent long-duration talking-face videos.

\section{Ethical Consideration}
\label{sec:ethical}
Our work focuses on enabling real-time and long-horizon audio-driven avatar generation, which naturally raises concerns related to privacy, consent, and potential misuse. All identity data used in training and evaluation is collected with permission, and our method does not store or reconstruct unauthorized identities. While high-fidelity avatars may be susceptible to impersonation risks, our system is intended solely for legitimate telepresence and interactive applications. We encourage responsible deployment practices such as access control and watermarking to prevent malicious use.

\end{document}